\RequirePackage{xcolor}
\documentclass[review]{elsarticle}
\usepackage[numbers]{natbib}
\usepackage{microtype}
\usepackage{graphicx}
\usepackage{subfigure}
\usepackage{booktabs} 
\usepackage{hyperref}

\usepackage{lineno}
\usepackage{hyperref}
\modulolinenumbers[5]

\journal{Pattern Recognition}

\usepackage{nicefrac} 
\usepackage{graphicx}	
\usepackage{todonotes}	
\usepackage{amsthm}	
\usepackage{tabularx}	
\usepackage{nicefrac}
\usepackage{mathtools}
\usepackage{appendix}	
\usepackage{xassoccnt}	
\usepackage{makecell}	
\usepackage{mathrsfs}	
\usepackage{amsfonts}
\usepackage{mathtools}
\usepackage{amsmath}
\usepackage{amssymb}
\usepackage{mathrsfs}
\usepackage{dblfloatfix}
\usepackage{adjustbox}
\usepackage[]{xcolor}
\usepackage{url}

\usepackage{amsmath, amssymb}	
\DeclareMathOperator*{\argmin}{\arg\!\min}	
\DeclareMathOperator*{\arginf}{\arg\!\inf}

\usepackage[acronyms]{glossaries}	
\newacronym{VA-GAN}{VA-GAN}{visual attribution generative adversarial network}	
\newacronym{VR-GAN}{VR-GAN}{visualization for regression with a generative adversarial network}	
\newacronym{PGD}{PGD}{projected gradient descent}	
\newacronym{PFT}{PFT}{pulmonary function test}	
\newacronym{COPD}{COPD}{chronic obstructive pulmonary disease}	
\newacronym{CXR}{CXR}{chest x-ray}	
\newacronym{GAN}{GAN}{generative adversarial network}	
\newacronym{PA}{PA}{posterior-anterior}
\newacronym{LLR}{LLR}{local linearity regularizer}	
\newacronym{LLM}{LLM}{local linearity measure}
\newacronym{cGAN}{cGAN}{conditional GAN}
\newacronym{IRB}{IRB}{Institutional Review Board}
\newacronym{STA}{STA}{Spatially Transformed Attack}

\usepackage{listofitems}
\def\listresetpaste{equation,theorem,lemma,footnote,thm,lemma1}
\newcommand\Copy[2]
  {  
  \renewcommand{\do}[1]{
  \expandafter\global\expandafter\edef\csname copy##1:#1\endcsname{\csname the##1\endcsname}
  }
  \readlist*\mylist{\listresetpaste}
  \foreachitem\x\in\mylist[]{\do{\x}}
  \expandafter\gdef\csname labeled:#1\endcsname{#2}
  #2}
\newcommand\Paste[1]
  {
    \readlist*\mylist{\listresetpaste}
    \renewcommand{\do}[1]{
    \expandafter\edef\csname paste##1\endcsname{\csname the##1\endcsname}
    \setcounter{##1}{\csname copy##1:#1\endcsname}
    }
  \foreachitem\x\in\mylist[]{\do{\x}}
  
  \csname labeled:#1\endcsname
      \renewcommand{\do}[1]{
    \setcounter{##1}{\csname paste##1\endcsname}
    }
    \foreachitem\x\in\mylist[]{\do{\x}}
  }

\newcommand{\beginsupplement}{%	
        \setcounter{table}{0}	
        \renewcommand{\thetable}{S\arabic{table}}%	
        \setcounter{figure}{0}	
        \renewcommand{\thefigure}{S\arabic{figure}}%	
        \setcounter{page}{1}	
        \setcounter{section}{1}
        \renewcommand{\thesection}{\Alph{section}}
     }	
	
\usepackage{stackengine}	
\usepackage{scalerel}	
	
\stackMath	
\def\dyhat{-.25ex}	
\newcommand\myhat[1]{\ThisStyle{%	
              \stackon[\dyhat]{\SavedStyle#1}	
                              {\SavedStyle\widehat{\phantom{#1}}}}}	
                              	
\newcolumntype{Y}{>{\centering\arraybackslash}X}
\newcolumntype{f}{>{\hsize=.15\hsize}X}	
\newcolumntype{s}{>{\hsize=.25\hsize}X}	
\newcolumntype{k}{>{\hsize=.6\hsize}Y}	
\newcolumntype{y}{>{\hsize=0.95\hsize}Y}	
\newcolumntype{z}{>{\hsize=.13\hsize}X}

\begin{document}
\newpageafter{author}
\begin{frontmatter}

\title{Quantifying the Preferential Direction of the Model Gradient in Adversarial Training With Projected Gradient Descent}

\author[sci]{Ricardo Bigolin Lanfredi\corref{mycorrespondingauthor}}
\cortext[mycorrespondingauthor]{Corresponding author}
\ead{ricbl@sci.utah.edu}

\author[rad]{Joyce D. Schroeder}
\ead{joyce.schroeder@hsc.utah.edu}

\author[sci]{Tolga Tasdizen}
\ead{tolga@sci.utah.edu}

\address[sci]{Scientific Computing and Imaging Institute, 72 S Central Campus Drive, Room 3750, Salt Lake City, UT 84112, USA}
\address[rad]{Department of Radiology and Imaging Sciences, University of Utah School of Medicine, 30 North 1900 East, Room 1A071, Salt Lake City, UT 84132, USA}

\begin{abstract}
Adversarial training, especially \gls*{PGD}, has proven to be a successful approach for improving robustness against adversarial attacks. After adversarial training, gradients of models with respect to their inputs have a preferential direction. However, the direction of alignment is not mathematically well established, making it difficult to evaluate quantitatively. We propose a novel definition of this direction as the direction of the vector pointing toward the closest point of the support of the closest inaccurate class in decision space. To evaluate the alignment with this direction after adversarial training, we apply a metric that uses generative adversarial networks to produce the smallest residual needed to change the class present in the image. We show that \gls*{PGD}-trained models have a higher alignment than the baseline according to our definition, that our metric presents higher alignment values than a competing metric formulation, and that enforcing this alignment increases the robustness of models.
\end{abstract}


\end{frontmatter}

% \linenumbers

\glsresetall
\section{Introduction}
Deep learning models have been shown to suffer from a lack of robustness against directed attacks that produce only small perturbations in the original input~\citep{advreview}. Several attacks and defenses of varying strengths have been proposed~\citep{advreview}. Defenses include, for instance, training using adversarial examples as samples~\citep{pgd} and regularizing gradients~\citep{gradientpenalty2}. One of the most successful defenses in terms of resisting new attacks~\citep{pgd_prove} is \gls*{PGD} training~\citep{pgd}.

\begin{figure}[tb]
\begin{center}
	\centerline{
	\includegraphics[width=1\columnwidth,]{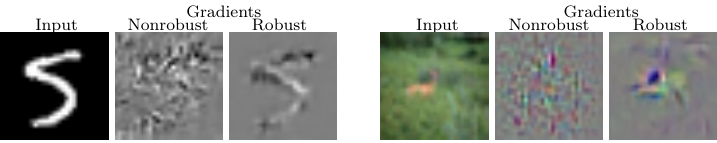}}
\caption{ Examples of the effect of robustness against adversarial attacks on the gradients of a model. Examples are from the MNIST dataset~\cite{mnist} (class 5) and the CIFAR-10 dataset~\cite{cifar10} (class deer). Robust models were trained using PGD~\cite{pgd}. The gradients of robust models are less noisy and have a higher magnitude in areas where the object of interest is located.} \label{fig:example_change}
\end{center}
\end{figure}

\Gls*{PGD} has been shown to change the gradient of the loss function of a trained model with respect to inputs $x$~\citep{interpretablepgd}. Examples of this gradient change can be seen in Figure \ref{fig:example_change}. Other robust training techniques also induce similar changes, modifying the gradient of the output class logits, including gradient regularization~\citep{alignment}. We focus our studies on \gls*{PGD} due to its success and widespread use~\citep{pgd_prove}, but our theoretical analysis is generic for any robust model. Only a few quantitative studies have related gradient direction to robustness. We propose a novel definition for this direction in classification problems. We formulate a gradient alignment metric $\alpha_{\Delta x}$ for a given sample $x$ as the expected cosine similarity between $\nabla_{\ell(x)}$ and $\Delta x$. We define $\nabla_{\ell(x)}$ as the gradient with respect to input $x$ of the function $\ell(x)$, related to the logits of a model and defined later in the paper, and $\Delta x$ as the vector pointing from $x$ to its closest neighbor $x^\prime$ in the support of class $\tilde{c}(x)$. The class $\tilde{c}(x)$ is the closest to $x$ in decision space that is not its ground truth class. 

We start by analyzing the robustness of models in a binary toy dataset, inspired by the work of \citet{spheres}, where the data for each class lies in one of two concentric spheres. We call this dataset \textit{Spheres}. The dataset was used by \citet{spheres} to demonstrate that nonzero generalization error can be the only cause of adversarial examples. Nonetheless, we use the dataset to prove a proportional relation between robustness and the proposed alignment metric. The theorems used for this proof are generic for multiclass classification problems, assuming local linearity and specific characteristics of the data distribution. Since $x^\prime$ is not straightforward to calculate for complex datasets, we also propose two methods to calculate an approximation $\hat{x}^\prime$ by applying generative adversarial training~\citep{gan}. We proceed to determine if the metric provides information for more complex datasets. Our results show that even though the most robust model did not always match the model with closest alignment, robust models trained with \gls*{PGD} have a larger average $\alpha_{\Delta x}$, and models trained to align their gradients with $\alpha_{\Delta x}$ have higher robustness. Furthermore, the proposed metric shows a closer alignment for robust models than an existing alignment metric proposed by \citet{alignment}.

\subsection{Related work}
\citet{alignment} mathematically defined the alignment of gradients after robust training as the alignment between the input image $x$ and the gradient of the logits of the output class $m(x)$ with respect to $x$, $\nabla\text{logit}(x)_{m(x)}$. Robustness is shown to be bounded by the sum of the given metric with other terms related to gradients and internal bias weights. However, the theoretical approach for the bias terms in the linear approximation of a model leads to a relatively loose bound in the relationship between robustness and alignment. We compare the metric proposed by \citet{alignment} against our metric and demonstrate that ours presents a closer alignment in practice. In the analyses of \citet{bugs} and \citet{boundary}, the gradient of robust models is shown to be better aligned with the vector connecting the centroid of two classes than the gradients of nonrobust models. However, this finding is restricted to binary linear models. The complex boundaries of piecewise linear deep learning models are unlikely to benefit from pointing in a single direction over the entire support of classes in high-dimensional datasets. Using local information of projection to the support of the opposite class, as we propose, is required to get flexible directions for each locally linear region of the model.

To study the gradients' alignment, we introduce a metric and propose to evaluate its correlation with robustness. Other metrics with similar motivations have been studied. The \gls*{LLM}~\citep{linearity}, for which a low value represents high local linearity of models around data points, is inversely correlated with the number of iterations in \gls*{PGD}. The metric was not evaluated directly against robustness, despite its potential. The CLEVER metric~\citep{clever} uses an estimated extreme value for the Lipschitz constant of the model to calculate a lower bound for robustness without having to perform evaluation attacks. This metric complements ours because it evaluates gradient magnitude instead of gradient direction. \citet{gradientleaking} proposed the gradient leaking metric, intuitively arguing that nonrobust models are caused by gradient components that are not aligned to the data manifold, assuming a continuous data manifold for all classes. The metric is defined as the sum of cosines from the angles between the model gradient and the components of a subspace calculated through PCA, which represents an estimation of the data manifold. Unlike ours, this metric does not establish a specific direction to which the gradient aligns but defines a set of possible directions. The assumptions made for the formulation of this metric do not generalize to datasets for which the gradient of robust models is perpendicular to the data manifold, such as the \textit{Spheres} dataset.

% method for estimating manifold: GAN vs PCA. 
% theoretical justification. 
% manifold vs continuous data manifold. 

We propose a penalty on the direction of $\nabla_{\ell(x)}$. This proposal adds to the literature of gradient penalties for robustness. Regularizing the gradient of a model with respect to its inputs has repeatedly been shown to increase its robustness~\citep{gradientpenalty2}. However, this penalty does not penalize the direction of the gradient. The \gls*{LLR}~\citep{linearity} was proposed and combined with \gls*{PGD} to allow faster training. Given that $\mathcal{L}(x)$ is the cross-entropy loss function, this penalty is equivalent to enforcing the $\nabla_{\mathcal{L}(x)}$ vector to be constant around data samples. However, the penalty enforces no direction. A penalty for aligning $\nabla\text{logit}(x)_{m(x)}$ with $x$ was proposed by \citet{alignment}, only as a future work. The alignment proposed by \citet{smoothalignment} enforces the proximity between a saliency map of the nonrobust model generated through a visual attribution technique and
$\nabla\text{logit}(x)_{y}$
, where $y$ is the ground truth class for $x$, with no theoretical justification for the choice of alignment direction. 

Other works have studied the relationship between a model's gradient and its robustness without directly studying the direction to which the gradient aligns. The method proposed by \citet{robustnesstransfer} transfers robustness from one model trained with \gls*{PGD} to other models by enforcing similar $\nabla_{\mathcal{L}(x)}$.
\citet{robustinterpretations} demonstrated that using their method to train models to have robust interpretation saliency maps indirectly leads to robust classification decisions. \citet{smoothnesslead} found that the gradient changes when performing robust training are a consequence of smoothness regularization over a model's decision functions in conjunction with the decision boundary orientation caused by conventional classification losses.

\section{Approach}
\subsection{Motivation and formulation of alignment metric}
\label{sec:mot}

The robustness of a decision model $m$ at a specific point $x$ can be defined as the signed distance of $x$ to the closest point where it is associated with a different model output. Formally, we define
\begin{equation}
\label{eq:robustness}
\rho(x) = \begin{cases}
    \,\,\,\,\;\inf \{\left\lVert \delta \right\rVert_{p}: m(x+\delta)\ne y\},\text{if } m(x)=y\\
    - \inf \{\left\lVert \delta \right\rVert_{p}: m(x+\delta)= y\},\text{if } m(x)\ne y
  \end{cases}
\end{equation}
where $\rho(x)$ is the robustness of $m$ against adversarial attacks at point $x$, $y$ is the ground truth class associated with $x$, $\text{inf}$ is the mathematical infimum, and $\left\lVert q \right\rVert_{p}$ is the p-norm of the vector $q$. We set a negative distance for misclassified examples to penalize errors and prevent models with a trivial decision boundary, i.e, one that always assigns the same class, to have infinite expected robustness. For this analysis, we will use the $L^2$ norm, i.e., $p=2$.

\begin{figure}[tb]
\begin{center}
	\centerline{
 
 \includegraphics[width=\columnwidth,trim={0cm 0.4cm 0cm 0cm},clip]{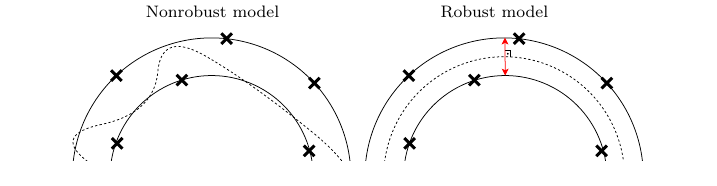}}
\caption{Illustration of slices of a subset of the \textit{Spheres} dataset. The concentric circles represent the manifolds of two opposite classes. The small $\times$s represent a few training examples from each class. The dashed lines represent the decision boundary of the two models. A nonrobust model is represented on the left, with a decision boundary that crosses the manifold of classes several times. A perfectly robust model is shown on the right, where the gradient of the decision function, i.e., the normal vector to the decision boundary, is locally aligned to the vector connecting both manifolds, represented by the red arrow.  } \label{fig:example_spheres}
\end{center}
\end{figure}

We analyze the \textit{Spheres} dataset proposed by \citet{spheres} to hypothesize about a specific aspect of robustness: the association between the robustness of a decision model and the direction of its logits' gradient with respect to inputs. We use this dataset because its simplicity allows for a more accessible analysis, whereas Deep ReLU networks can still fail in modeling it robustly~\cite{spheres}. The \textit{Spheres} dataset is composed of two classes with support on the surface of two hyperspheres, of radius 1.0 (class -1) and 1.3 (class 1), in a 500-dimensional space. The prior probability for each class is $0.5$, and the distribution of samples is uniform on each hypersphere surface. An illustration of how the robustness of a model may affect its gradient for this dataset is presented in Figure \ref{fig:example_spheres}.
We use the term support of a class $c$ as defined by $supp_c=\{x\in X \mid P(x|y=c)>0\}$, where $X$ is the domain of inputs, and $P(q|u)$ is the probability of event $q$ conditioned on the occurrence of event $u$.

For this dataset, the optimally robust model $m_p$ has its decision rule defined by 
\begin{equation}
c = \begin{cases}
    \,\,\,\;1, & \text{if } \left\lVert x \right\rVert_2 > t\\
    -1, & \text{if } \left\lVert x \right\rVert_2 < t
  \end{cases}, t = 1.15.
\end{equation}
This model has a margin of 0.15 between the decision boundary and any data point. Note that the expected robustness $\mathbb{E}_{x\sim X}\left[\rho(x)\right] $ is the same for all values of $t$, and the decision to choose $t=1.15$ is based on classification margins. A differentiable decision model can be obtained by defining $ \text{logit}(x)_1 = \left\lVert x \right\rVert_2-1.15$ and $ \text{logit}(x)_{-1} = -\left\lVert x \right\rVert_2+1.15$, where $\text{logit}(x)_c$ is the logit of a model for class $c$ and input $x$. If we define a loss for the model as $-y\times\text{logit}(x)_1+y\times\text{logit}(x)_{-1}$, this model has radial gradients with respect to the input. The gradients of the defined loss point toward the origin for class 1 and away from the origin for class -1. We note that the optimally robust model has gradients that point from the support of one class to the closest point $x^\prime$ of the support of the other class. We denote the vector connecting $x$ to $x^\prime$ as $\Delta x$. We will proceed to theoretically justify the importance of $\Delta x$.

\newtheorem{lemma1}{Lemma}
\newtheorem{lem}[lemma1]{Lemma}
\newtheorem{thm}{Theorem}
\begin{lem}
\label{lemma:locallinearity}
Let $R(x)\coloneqq\sup \{r\in\mathbb{R}\text{, given that } \text{ for all } v\in\mathbb{R}^n \text{ such that }\\ \left\lVert x - v \right\rVert_2<r, \text{there is a } W\in\mathbb{R}^{n\times C}\text{ and there is a } b\in\mathbb{R}^{C} \text{ such that } \text{logit}(v)=W^{T}v +b\}$, where $\text{logit}(v)\in \mathbb{R}^C$ is a vector representing all logits of a decision model $m$ for input $v$, $x\in \mathbb{R}^n$, $C$ is the total number of classes of a dataset, and $\text{sup}$ is the mathematical supremum. In other words, $R(x)$ is the radius of the largest hypersphere around $x$ where a decision model $m$ can be defined by a specific linear decision function. Assuming that $R(x)\ge\left\lvert\rho(x)\right\rvert$ for $m$, then the logits of $m$ can be modeled as a linear function, i.e, $\text{logit}(x)=W^{T}v +b$, for robustness assessment, resulting in a robustness magnitude at $x$ given by
\begin{equation}
\label{eq:robustnesslemma1}
\begin{aligned}
\left\lvert\rho(x)\right\rvert=\min_{c}\left\lvert\rho(x)_c\right\rvert=\left\lvert\rho(x)_{\tilde{c}(x)}\right\rvert,
\end{aligned}
\end{equation}
where
\begin{equation}
\label{eq:linearlemma1}
\begin{aligned}
\left\lvert\rho(x)_c\right\rvert={\left\lvert\frac{\text{logit}(x)_{y}-\text{logit}(x)_{c}}{\left\lVert W_{:,y}-W_{:,{c}} \right\rVert_2}\right\rvert},
\end{aligned}
\end{equation}
\begin{equation}
\label{eq:clemma1}
\begin{aligned}
\tilde{c}(x)\coloneqq\argmin_{c\ne y}\{\inf\{\left\lVert v\right\rVert_2:m(x+v)=c\} \}=\\=\argmin_{c\ne y}{\left\lvert\rho(x)_c\right\rvert}=\argmin_{c\ne y}{\left\lvert\frac{\text{logit}(x)_{y}-\text{logit}(x)_{c}}{\left\lVert W_{:,y}-W_{:,{c}}
%TODO: modify {c}(x) to c
\right\rVert_2}\right\rvert},
\end{aligned}
\end{equation}
and $W_{:,j}$ is the the $j^{th}$ column of the $W$ matrix.
\end{lem}

Lemma \ref{lemma:locallinearity} follows from the Lemma 1 proposed and proved by \citet{alignment}. They also empirically showed that, despite $R(x)\ge\left\lvert\rho(x)\right\rvert$ not holding, the linearization is a good approximation. We will refer to models satisfying the assumption in Lemma \ref{lemma:locallinearity} as being locally linear around $x$.

\Copy{th1}{
\begin{thm}
\label{theorem:pair}
Let $\textit{sim}(u,v)$ be the alignment between vectors $u$ and $v$, defined by their cosine similarity $\textit{sim}(u,v)=\frac{\langle u,v\rangle}{\left\lVert u \right\rVert_2\left\lVert v \right\rVert_2}$, where $\langle u,v\rangle$ is the dot product between $u$ and $v$. Let $m$ be a classification model and let $c^*(x) \coloneqq \argmin_{c\ne m(x)}\{\inf\{\left\lVert v\right\rVert_2:m(x+v)=c\} \}$, where $m(x)$ is the class outputted by model $m$ when its input is $x$. In other words, $c^*(x)$ is the closest class to $x$ in decision space that is not the output of model $m$. For a pair of input examples $x_i$ and $x_j$, of different classes $i$ and $j$, respectively, around which $m$ is locally linear and for which\footnote{The notation of these two equations uses sets in which $\{a,b\}=\{i,j\}$ means that either $a=i$ and $b=j$ or $a=j$ and $b=i$.} $\{c^*(x_i),m(x_i)\}=\{i,j\}$ and $\{c^*(x_j),m(x_j)\}=\{i,j\}$, i.e., for which $j$ and $i$ are the two closest classes to both $x_i$ and $x_j$, the combined robustness $\rho(x_i)+\rho(x_j)$ of $m$ is directly proportional to $\alpha$ according to $\rho(x_i) + \rho(x_j) = \left\lVert x_j-x_i \right\rVert_2\times\alpha$, where  $\alpha=\textit{sim}(x_j-x_i, \nabla{\text{logit}(x_i)_j}-\nabla{\text{logit}(x_i)_i})=\textit{sim}(x_i-x_j, \nabla{\text{logit}(x_j)_i}-\nabla{\text{logit}(x_j)_j})$.
\end{thm}
}

\begin{figure}[tb]
\begin{center}
	\centerline{\includegraphics[width=0.415\columnwidth,]{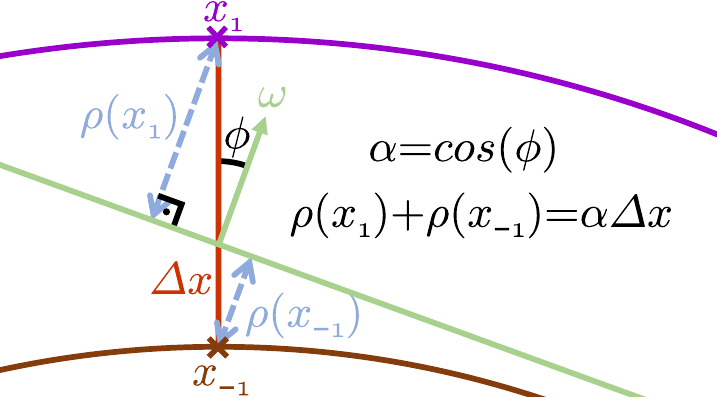}}
\caption{Illustration of the result from Theorem \ref{theorem:pair} for the \textit{Spheres} dataset. The green straight line represents the decision boundary of an arbitrary classifier, and the purple top and brown bottom arcs represent a section of the support of classes 1 and -1, respectively. The quantity $\alpha$, defined as the cosine of the angle $\phi$ between the normal of the decision boundary $\omega$ and the vector $\Delta x$ connecting both input samples $x_1$ and $x_{-1}$, is proportional to the sum of robustness of both input samples, $\rho(x_1)$ and $\rho(x_{-1})$. } \label{fig:th1}
\end{center}
\end{figure}
Theorem \ref{theorem:pair} establishes that, given assumptions of local linearity and symmetry of the closest decision boundaries, the sum of the robustness of two inputs, for which either the model's output or the closest class is the ground truth for that example, is proportional to the alignment between the model gradient and a vector connecting both inputs. The proof for Theorem \ref{theorem:pair} is in Section \ref*{sec:proofs} in the Supplementary Material. An illustration of the theorem claim for the \textit{Spheres} dataset is given in Figure \ref{fig:th1}. 

For use of Theorem \ref{theorem:pair} in Theorem \ref{theorem:average}, we define 
\begin{equation}
\ell(x)\coloneqq\text{logit}(x)_{\tilde{c}(x)}-\text{logit}(x)_{y},
\end{equation}
leading to $\alpha=\textit{sim}(x_{\tilde{c}(x)}-x_y, \nabla_{\ell(x)})$. 
We proceed to formulate a global metric to measure the robustness of a given model, where the robustness is the expected sample robustness with respect to the data distribution.

\Copy{th2}{
\begin{thm}
\label{theorem:average}
Assuming that, for a multiclass dataset of classes $\mathcal{C}$, 
\begin{enumerate}
\item it is possible to define $K$ mutually exclusive sets $\mathscr{S}_k$, each containing regions of the supports of two classes $i_k$ and $j_k$, where $\bigcup\limits_{c\in \mathcal{C}}supp_c = \bigcup\limits_{k=1}^{K} \mathscr{S}_k$, i.e., the $K$ sets cover the whole space of the support of classes;
\item for each $\mathscr{S}_k$, it is possible to define a bijection between the respective regions of support of classes $i_k$ and $j_k$ such that, given all bijection pairs $(x_{i_k},x_{j_k})$, $x_{i_k}\in supp_{i_k}$ and $x_{j_k}\in supp_{j_k}$,
\begin{enumerate}
\item $P(x_{i_k})=P(x_{j_k})$, where $P(q)$ is the probability of sampling %q$,
\item a decision model $m$ is locally linear around $x_{i_k}$ and $x_{j_k}$;
\item ${\{c^{*}(x_{i_k}), m(x_{i_k})\}=\{i_k,j_k\}}$ and ${\{c^{*}(x_{j_k}), m(x_{j_k})\}=\{i_k,j_k\}}$ ;
\end{enumerate}
\end{enumerate}
then the expected robustness of $m$, $\rho_m$,  is related to the expected alignment $\overline{\alpha}\,$ between 
$\nabla_{\ell(x)}$ and $\Delta x$ of pairs $(x_{i_k},x_{j_k})$ over all $\mathscr{S}_k$, according to
\begin{equation}
\label{eq:bounds}
\rho_m \ge \frac{\inf(\mathscr{D}) \times \overline{\alpha}}{2} \,\,\text{ and }\,\, \overline{\alpha}\ge \frac{2\times\rho_m}{\sup(\mathscr{D})},
\end{equation}
where $\mathscr{D}$ is the set of distances $\left\lVert x_{i_k}-x_{j_k} \right\rVert_2$ over all pairs $(x_{i_k},x_{j_k})$ over all $\mathscr{S}_k$.
\end{thm}
}

Theorem \ref{theorem:average} sets bounds for the relationship between the expected robustness of a model and the average alignment between the model's gradient and vectors connecting the inputs of two adjacent classes. The proof for Theorem \ref{theorem:average} is given in Section \ref*{sec:proofs} in the Supplementary Material. The local linearity assumption of Theorem \ref{theorem:average} is likely to hold only if the bijection can be established between the closest points of the supports of adjacent classes. Therefore, we define a metric using the concept of vector $\Delta x$ pointing to the closest point of the support of $\tilde{c}(x)$. This metric is given by
\begin{equation}
\label{eq:metric}
\begin{aligned}
\overline{\alpha_{\Delta x}} = \int{P(x)\frac{\langle \Delta x,\nabla_{\ell(x)}\rangle }{\left\lVert \Delta x \right\rVert_2 \left\lVert   \nabla_{\ell(x)} \right\rVert_2}dx},\\
\Delta x = \arginf_{r} \{\left\lVert r \right\rVert_2:x+r\in supp_j, x\in supp_y, j\neq y \}, j = \tilde{c}(x)  .
\end{aligned}
\end{equation}
In practice, $\tilde{c}(x)$ can be calculated using the linear approximation given in (\ref{eq:clemma1}), and we propose methods for calculating $\Delta x$ in Section \ref{sec:gendeltax}. Given that assumptions from Theorem \ref{theorem:average} are satisfied for pairs of closest points, $\overline{\alpha_{\Delta x}}$ as defined (\ref{eq:metric}) will be equal to $\overline{\alpha}$ in (\ref{eq:bounds}). For the \textit{Spheres} dataset, it is possible to establish a bijection as required by Theorem \ref{theorem:average} using points of opposite classes along the same radial direction. Since the prior probability of both classes is balanced, and the probability distribution in both supports is uniform, $P(x_{-1})=P(x_1)$ holds for any pair of points. The assumption $\{c^*(x_i),m(x_i)\}=\{i,j\}$ and $\{c^*(x_j),m(x_j)\}=\{i,j\}$ ($i\ne j$) always holds for binary datasets. Thus, except for a possible violation of the local linearity assumption, Theorem \ref{theorem:average} holds for the \textit{Spheres} dataset. Additionally, the distance between closest points is constant, so both bounds can be combined into an equality $\rho_{m}= \left\lVert\Delta x\right\rVert_2 \times \overline{\alpha_{\Delta x}}/2$. According to Theorem \ref{theorem:average}, the optimally robust model has  $\rho_{m}=(0.3\times 1)/2 = 0.15$, which is the expected value. However, the assumptions needed for applying Theorem \ref{theorem:average} are unlikely to hold exactly for more complex datasets. We perform empirical analysis to evaluate the alignments of such datasets in Section \ref{sec:aer}. Section \ref{sec:linearity} provides an additional survey of the approximation defined in Lemma \ref{lemma:locallinearity} for the models tested in Section \ref{sec:aer}.

 \subsection{Gradient penalty} 
 We propose to test if steering $\nabla_{\ell(x)}$ by adding a penalty $L_\alpha$ to a supervised classification loss will increase the robustness of a model. This penalty is given by   
\begin{equation}
\label{eq:penalty}
L_\alpha\coloneqq-\lambda_\alpha  \overline{\alpha_{\Delta x}}, L=L_\alpha+\mathcal{L}
\end{equation}
where $\lambda_\alpha$ is a hyperparameter controling the importance of loss $L_\alpha$ over total loss $L$, and $\mathcal{L}$ is the cross-entropy loss function. This penalty is not meant to replace other robust training methods, but to show that increasing alignment increases robustness.
\subsection{Estimating $\Delta x$}
\label{sec:gendeltax}
For almost all real-world datasets, $\Delta x$ is not trivial to find. To generate it, we use generative adversarial training~\citep{gan} to characterize the support of classes.
\subsubsection{Direct generation of $\widehat{\Delta x}$}
\label{sec:vrgan}
\begin{figure*}[t]
\begin{center}
\centerline{\includegraphics[width=0.85\textwidth,clip]{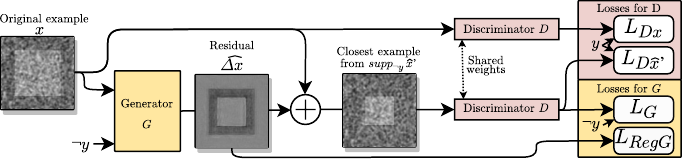}}
\caption{Training diagram for directly generating residual approximations $\widehat{\Delta x}$. The terms $L_{Dx}$, $L_{D\hat{x}}$, and $L_{G}$ are classification losses, and $L_{RegG}$ penalizes the length of $\widehat{\Delta x}$.} \label{fig:vrgan}
\end{center}
\end{figure*}
We train a generator $G$ to directly produce the residuals needed to convert from one class to another. The formulation draws from the VR-GAN method~\citep{vrgan}, modified to work with classification tasks instead of regression tasks. For practical reasons, we use this formulation only on binary datasets. Figure \ref{fig:vrgan} shows the overall formulation for training. 
Residuals $\widehat{\Delta x}$ are produced according to
\begin{equation}
\widehat{\Delta x}=G(x, \neg {y}),\hat{x}^\prime = x + \widehat{\Delta x}
\end{equation}
where $\neg{y}$ is the opposite class to ground truth $y$, and $\hat{x}^\prime$ is the approximated example of $\neg {y}$ closest to $x$. We set up an adversarial loss given by
\begin{equation}
\begin{aligned}
L_{Dx} = \mathbb{E}\left[\mathcal{L}(D(x), y)\right], L_{D\hat{x}^\prime}= \mathbb{E}\left[\mathcal{L}(D(\hat{x}^\prime), y)\right],\\L_G = \mathbb{E}\left[\mathcal{L}(D(\hat{x}^\prime),\neg {y})\right],
\end{aligned}
\end{equation}
where $\mathcal{L}$ is binary cross-entropy, $L_{Dx}$ and $L_{D\hat{x}^\prime}$ are losses for which a discriminator $D$ is optimized, and $L_G$ optimizes a generator $G$. $G$ is trained to fool $D$ ($L_G$), whereas $D$ is trained not to be fooled ($L_{D\hat{x}^\prime}$). This adversarial setup, combined with a traditional supervised loss ($L_{Dx}$), should make $D$ accept only $\hat{x}^\prime$ that are in $supp_y$. The loss $L_{D\hat{x}^\prime}$ should have a smaller weight than $L_{Dx}$ so that, if $G$ is generating perfect modifications, $D$ can still learn $supp_y$. Finally, we define the term 
\begin{equation}
L_{RegG} = \frac{\left\lVert \widehat{\Delta x} \right\rVert_2}{\sqrt{n}},
\end{equation}
where $n$ is the dimensionality of $x$. This penalty is used to enforce that $\hat{x}^\prime$ is the closest point in the learned support of classes. The final optimization is given by
\begin{equation}
\begin{aligned}
G^* = \argmin_{G} (\lambda_{G} L_{G}+\lambda_{RegG} L_{RegG} ),\\D^* = \argmin_{D} (\lambda_{Dx} L_{Dx}+, \lambda_{D\hat{x}^\prime} L_{D\hat{x}^\prime}),
\end{aligned}
\end{equation}
\noindent where $\lambda_{G}$, $\lambda_{RegG}$, $\lambda_{Dx}$, $\lambda_{D\hat{x}^\prime}$ are hyperparameters controlling the relative importance of each loss term.
\subsubsection{Indirect generation of $\widehat{\Delta x}$}
\label{sec:defensegan}
\begin{figure}[t]
\begin{center}
\centerline{\includegraphics[width=0.45\columnwidth,clip]{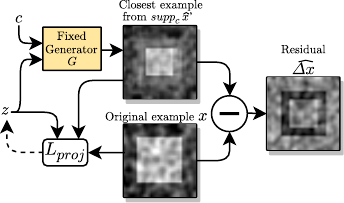}}
\caption{Diagram for indirectly generating $\widehat{\Delta x}$ connecting an input example $x$ to its estimated closest example $\hat{x}^\prime$ from class $c$. The dashed line depicts the feedback from loss $L_{proj}$ to iteratively update latent space vector $z$ .} \label{fig:cgan}
\end{center}
\end{figure}
We propose an indirect method for estimating $\Delta x$ as a strategy to scale it to multiclass datasets. This method can be easily adapted to new datasets since it uses, with no modifications, any available established \gls*{cGAN}~\cite{cgan} for the datasets of interest. After training a conditional generator $G$, we follow a similar algorithm for projection to the \gls*{GAN} manifold as in the Defense-GAN method~\cite{defensegan}. In other words, we iteratively find, for each image $x$, the optimal latent space vector 
\begin{equation}
z^*_{c}=\argmin_{z}{L_{proj}(x,z,c)},
\end{equation}
where $c$ is one of the dataset classes, and $L_{proj}$ is a loss that measures both the distance between input image $x$ and $G(z,c)$, and the likelihood of $z$, given by
\begin{equation}
\label{eq:lproj}
L_{proj}(x,z,c) = \frac{\left\lVert x-G(z,c) \right\rVert_2^2}{n} + \lambda_{Reg\hspace{0.02cm}z} \frac{\left\lVert z \right\rVert_2^2}{n},
\end{equation}
where $\lambda_{Reg\hspace{0.02cm}z}$ is a hyperparameter controlling the relative importance of the likelihood of latent space vector $z$. The result of the projection, $G(z^*_c,c)$, is an estimation of the example $x^\prime$ for class $c$ that is closest to $x$. We can then indirectly calculate
\begin{equation}
\widehat{\Delta x} = G(z^*_{c},\tilde{c}(x)) - x.
\end{equation}
 We use a penalty on the norm of $z$ because a Gaussian prior was used when randomly sampling $z$ during training for the chosen \glspl*{cGAN}. A representation of the algorithm can be found in Figure \ref{fig:cgan}. More details of the optimization process can be found in Section \ref{sec:cgan} in the Supplementary Material.

\subsection{Adversarial defense and attack}

Adversarial training with \gls*{PGD}~\citep{pgd} aims to find a robust parameterized classifier $m$ by optimizing 
\begin{equation}
\min_{m}\max_{\delta}\mathcal{L}(m(x+\delta),y), \left\lVert \delta \right\rVert_{p} < \epsilon,
\end{equation}
where $\delta$ is a residual with a limited norm, $\mathcal{L}$ is a classification loss function, $x$ is an input example, $y$ is the ground truth label for input $x$, $\epsilon$ is a hyperparameter defining the desired robustness distance, and, commonly, $p=\infty$. The inner maximization is performed using iterative gradient ascent with $\kappa$ steps of size $\eta$. Data projections are performed after each step to satisfy norm limits and data intensity ranges. The method can also be used as a strong adversarial attack.

\section{Experiments}

We empirically analyzed\footnote{Code is available at \url{https://github.com/ricbl/gradient-direction-of-robust-models}
} if improving the robustness of a model using \gls*{PGD} training increased its alignment $\overline{\alpha_{\Delta x}}$, and if increasing $\overline{\alpha_{\Delta x}}$ by using the alignment penalty $L_\alpha$ improved robustness. We compared both training methods against a baseline using plain supervised learning. We also compared the values given by our metric $\overline{\alpha_{\Delta x}}$ against values given by the metric proposed by \citet{alignment}, to which we refer as $\overline{\alpha_x}$. We adapted the metric, adding a normalization by  $\left\lVert x \right\rVert_2$ to change the range of values to $[0,1]$, to allow a comparison with $\alpha_{\Delta x}$. The metric was modified as
\begin{equation}
\overline{\alpha_{x}} = \int{P(x)\frac{\left\lvert\langle x,\nabla\text{logit}(x)_{m(x)} \rangle \right\rvert}{\,\,\left\lVert x \right\rVert_2 \left\lVert   \nabla\text{logit}(x)_{m(x)} \right\rVert_2} dx},
\end{equation}
where $m(x)$ is the class outputted by the model and $\left\lvert q \right\rvert$ is the absolute value of $q$. We performed five experiments for models we trained. We report the average resulting values and their standard deviations. Section \ref*{sec:setup} of the Supplementary Material presents details about the experimental setup and hyperparameters.
\subsection{Datasets}
We performed most evaluations on six datasets, two of which were synthetic datasets for which we could define the correct $\Delta x$. For the \textit{Spheres} dataset, defined in Section \ref{sec:mot}, samples were always drawn randomly at runtime from a standard Gaussian distribution and normalized to the radius of the respective class. The correct $\Delta x$ was calculated by
\begin{equation}
\Delta x = \begin{cases}
    \,\,\,\:0.3\,\; x\,\,, & \text{if } \left\lVert x \right\rVert_2=1\\
    -0.3 \frac{x}{1.3}, & \text{if } \left\lVert x \right\rVert_2=1.3
  \end{cases}.
\end{equation}
We created another synthetic dataset, which we refer to as \textit{Squares}, composed of images with $224\times224$ pixels of centered squares with sides of 142 or 88 pixels. To make the images unique, spatially smoothed Gaussian noise was randomly sampled for each image and added to it. The direction of alignment $\Delta x$ for this dataset was calculated using the subtraction of noiseless images from each class. 

To evaluate with more complex datasets, we used the MNIST dataset~\citep{mnist}, containing handwritten digits; the CIFAR-10 dataset~\cite{cifar10}, containing low-resolution natural images; and a binary \gls*{CXR} dataset~\cite{copddataset}, which we refer to as \textit{COPD}. For the MNIST and CIFAR-10 datasets, a fixed set of 10\% of the training set was used for validation. To enable the comparison of the method in a multiclass setting with a corresponding binary setting, we also tested the method with a binary MNIST dataset selecting only two similar digits, 3 and 5, which we refer to as MNIST-3/5. The \textit{COPD} dataset contained \gls*{PA} \glspl*{CXR} labeled for \gls*{COPD} using \glspl*{PFT} and was adopted with \gls*{IRB} approval\footnote{IRB\_00104019, PI: Schroeder MD}. The intensity range of all image datasets was adjusted to $[-1,1]$. For \textit{COPD} and MNIST-3/5, the method presented in Section \ref{sec:vrgan} was used to estimate $\Delta x$, whereas for MNIST and CIFAR-10, the method presented in Section \ref{sec:defensegan} was used. More details about the datasets are given in the Supplementary Material, and example images can be seen in Figure \ref{fig:gradients} and Figure \ref{fig:gradientsGAN}. We also included a limited evaluation of the metrics with the ImageNet dataset~\cite{imagenet}. 

\subsection{Validating the estimation of $\Delta x$}

\begin{figure*}[!ht]
\begin{center}
\centerline{\includegraphics[width=0.97\linewidth]{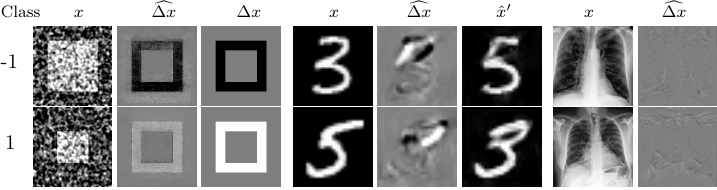}}
\caption{Results of the generated $ \widehat{\Delta x} $ through the direct method for random samples in the image datasets. The $x^\prime$ column is suppressed for the \textit{COPD} dataset because of its resemblance to $x$.} \label{fig:gradients}
\end{center}
\end{figure*}
\begin{figure*}[!ht]
\begin{center}
\centerline{\includegraphics[width=0.97\linewidth]{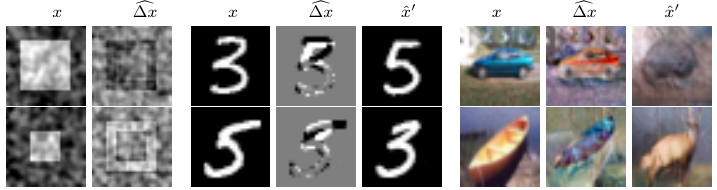}}
\caption{Results of the generated $ \widehat{\Delta x} $ through the indirect method for random samples in the image datasets. The classes for the CIFAR-10 dataset were chosen at random, and the images represent a transition from car to frog and from boat to deer. } \label{fig:gradientsGAN}
\end{center}
\end{figure*}
\begin{table}[t]

\begin{center}
\caption{Examples of measures for generated $ \widehat{\Delta x} $ for the \textit{Spheres} datasets for two random samples.} \label{table:gradients}
\begin{small}
\begin{tabular}{@{}lccccc@{}}
\toprule	Class & $\left\lVert x \right\rVert_2$             & $\left\lVert \widehat{\Delta x} \right\rVert_2$ & $\left\lVert \widehat{x}^\prime \right\rVert_2$ & $\left\lVert x^\prime \right\rVert_2$ & $\textit{sim}(\Delta x,\widehat{\Delta x})$ \\						\midrule

	\multicolumn{1}{l}{-1}	& 1.0& .307 & 1.27	& 1.3 & .858	\\
	\multicolumn{1}{l}{\hspace{0.123cm}1}	& 1.3	& .360 & 0.99 & 1.0	& .898	\\	\bottomrule

\end{tabular}
\end{small}
\end{center}

\end{table}
To validate the direct method for estimating $\Delta x$, we applied it to both datasets for which we know the correct $\Delta x$ and measured the alignment between $\Delta x$ and $\myhat{\Delta x}$. We found that, for the \textit{Spheres} dataset, $\textit{sim}(\Delta x,\myhat{\Delta x})=0.874\pm0.019$, and, for the \textit{Squares} dataset, $\textit{sim}(\Delta x,\myhat{\Delta x})=0.893\pm0.058$, demonstrating close alignment. Table \ref{table:gradients} shows examples of results for the \textit{Spheres} dataset. Figure \ref{fig:gradients} shows examples of generated $\myhat{\Delta x}$. The generated $\myhat{\Delta x}$ were similar to the expected $\Delta x$ for the \textit{Spheres} and the \textit{Squares} datasets. 
For the \textit{COPD} dataset, $\myhat{\Delta x}$ had small norms, with changes mainly around the diaphragms and the upper lungs. Diaphragm shape and position are used as \gls*{COPD} evidence in \gls*{CXR}~\citep{copdxray}. The small norms are likely due to the continuous characteristic of disease severity, which leads to the support of both classes being on the same manifold. Furthermore, most samples had \gls*{PFT} values near the threshold between classes.

For the indirect method for estimating $\Delta x$, the chosen \gls*{cGAN} algorithm produced visually good results for the MNIST dataset without any changes to the hyperparameters used for CIFAR-10. We did not apply the method to the \textit{Spheres} dataset because of incompatibilities with image-oriented \glspl*{cGAN}. To quantitatively validate the indirect method, we applied it to an adapted version of the \textit{Squares} dataset, \textit{Squares32}, for which images had size $32\times32$, finding an alignment of $\textit{sim}(\Delta x,\myhat{\Delta x})=0.661\pm0.070$. This alignment is good but smaller than the alignments found with the direct method. Visually, we found a worse representation of the variability in the background of images for this method when compared to the method proposed in Section \ref{sec:vrgan}, which needs to generate only the differences between classes, adopting the background from the original image. 
Examples of the generated images can be seen in Figure \ref{fig:gradientsGAN} and Section \ref*{sec:severalganimages} of the Supplementary Material. A comparison between the generated images for MNIST-3/5 and MNIST showed that the indirect method generates sharper and slightly more distant images. For the CIFAR-10 dataset, the generated images resembled the destination class in most cases but failed more often with some destination classes, such as airplanes, and with images containing certain types of scenes, such as white backgrounds.
The generated images for MNIST almost always resembled the destination class.

\subsection{Alignment and robustness}
\label{sec:aer}
\begin{table*}[tbh!]
\setlength\tabcolsep{2.7pt}

\begin{center}
\caption{Results for robustness, alignments, and accuracy after training using a plain supervised training baseline (B), alignment penalty ($L_{\alpha}$), and adversarial training with \gls*{PGD} (P) for six datasets: \textit{Spheres} (S), \textit{Squares} (Q), MNIST-3/5 ($\mu$), \textit{COPD} (X), MNIST (M), CIFAR-10 (C). Attacks employed to calculate robustness include PGD~\citep{pgd}, with two types of norms, and the black-box Square Attack~\citep{blbox} (BlBox).} \label{table:results}
\begin{small}
\begin{tabular}{@{}ccccccc@{}}
\toprule	Setup & \makecell {Accuracy \\(\%)}  & \makecell{$\epsilon_{50\%}$\\ PGD$_{p=\infty}$} & \makecell{$\epsilon_{50\%}$\\ PGD$_{p=2}$} & \makecell{$\epsilon_{50\%}$\\ BlBox$_{p=\infty}$}  &   \makecell{$\overline{\alpha_{\Delta x}}$ \\ (ours)}   &   $\overline{\alpha_{x}}$     \\						\midrule

\multicolumn{1}{l}{S-B}	& 99.4$\pm$0.8	& 0.0058$\pm$.0001	& .099$\pm$.009	& 0.0063$\pm$.0004	& .659$\pm$.028	& .659$\pm$.028	\\	
\multicolumn{1}{l}{S-$L_{\alpha}$}	& 100.$\pm$0.0	& 0.0076$\pm$.0002	& .133$\pm$.001	& 0.0083$\pm$.0003	& .886$\pm$.002	& .886$\pm$.002	\\	
\multicolumn{1}{l}{S-P}	& 100.$\pm$0.0	& 0.0073$\pm$.0001	& .126$\pm$.002	& 0.0080$\pm$.0001	& .851$\pm$.001	& .851$\pm$.001	\\	\midrule
\multicolumn{1}{l}{Q-B}	& 100.$\pm$0.0	& 0.031$\pm$.008	& 16.5$\pm$1.6	& 0.170$\pm$.041	& .026$\pm$.007	& .022$\pm$.006	\\	
\multicolumn{1}{l}{Q-$L_{\alpha}$}	& 100.$\pm$0.0	& 0.435$\pm$.061	& 61.9$\pm$5.4	& 0.405$\pm$.001	& .926$\pm$.004	& .357$\pm$.009	\\	
\multicolumn{1}{l}{Q-P}	& 100.$\pm$0.0	& 0.501$\pm$.030	& 61.3$\pm$2.0	& 0.337$\pm$.011	& .222$\pm$.042	& .151$\pm$.051	\\	\midrule
\multicolumn{1}{l}{$\mu$-B}	& 99.3$\pm$0.4	& 0.198$\pm$.012	& 2.36$\pm$.11	& 0.232$\pm$.010	& .171$\pm$.029	& .013$\pm$.003	\\	
\multicolumn{1}{l}{$\mu$-$L_{\alpha}$}	& 99.5$\pm$0.5	& 0.357$\pm$.011	& 3.88$\pm$.08	& 0.351$\pm$.009	& .678$\pm$.078	& .196$\pm$.010	\\	
\multicolumn{1}{l}{$\mu$-P}	& 99.5$\pm$0.2	& 0.547$\pm$.007	& 4.20$\pm$.15	& 0.495$\pm$.005	& .345$\pm$.018	& .040$\pm$.032	\\	\midrule
\multicolumn{1}{l}{X-B}	& 66.3$\pm$1.5	& 0.006$\pm$.0021	& 0.73$\pm$.33	& 0.021$\pm$.0063	& .016$\pm$.005	& .002$\pm$.0003	\\	
\multicolumn{1}{l}{X-$L_{\alpha}$}	& 64.2$\pm$5.4	& 0.020$\pm$.0047	& 2.91$\pm$.65	& 0.028$\pm$.0074	& .163$\pm$.009	& .064$\pm$.019	\\	
\multicolumn{1}{l}{X-P}	& 64.7$\pm$2.7	& 0.063$\pm$.0294	& 4.97$\pm$2.4	& 0.072$\pm$.0324	& .081$\pm$.007	& .023$\pm$.013	\\	\midrule
\multicolumn{1}{l}{M-B}	&	99.2	$\pm$	0.2	&	0.171	$\pm$	.004	&	2.19	$\pm$	.06	&	0.188	$\pm$	0.003	&	.075	$\pm$	.006	&	.038	$\pm$	.005	\\	
\multicolumn{1}{l}{M-$L_{\alpha}$}	&	99.2	$\pm$	0.1	&	0.325	$\pm$	.009	&	3.55	$\pm$	.06	&	0.308	$\pm$	0.007	&	.592	$\pm$	.009	&	.320	$\pm$	.009	\\	
\multicolumn{1}{l}{M-P}	&	99.4	$\pm$	0.0	&	0.554	$\pm$	.007	&	4.32	$\pm$	.07	&	0.489	$\pm$	0.001	&	.181	$\pm$	.012	&	.043	$\pm$	.008	\\	\midrule
\multicolumn{1}{l}{C-B}	&	84.0	$\pm$	0.2	&	0.008	$\pm$	.0003	&	.23	$\pm$	.121	&	0.008	$\pm$	0.005	&	.008	$\pm$	.003	&	.008	$\pm$	.000	\\	
\multicolumn{1}{l}{C-$L_{\alpha}$}	&	81.1	$\pm$	0.6	&	0.012	$\pm$	.0006	&	.44	$\pm$	.023	&	0.016	$\pm$	0.001	&	.025	$\pm$	.002	&	.020	$\pm$	.002	\\	
\multicolumn{1}{l}{C-P}	&	82.6	$\pm$	0.1	&	0.023	$\pm$	.0001	&	.77	$\pm$	.003	&	0.028	$\pm$	0.000	&	.041	$\pm$	.001	&	.031	$\pm$	.001	\\	\bottomrule
\end{tabular}
\end{small}
\end{center}

\end{table*}

\begin{table*}[tbh!]
\setlength\tabcolsep{2.7pt}

\begin{center}
\caption{Robustness against the Spatially Transform Attack~\cite{sta}, measured as the value of $\tau$ for which 50\% of examples were misclassified, for three training methods and five image datasets.} \label{table:resultssta}
\begin{small}
\begin{tabular}{@{}cccccc@{}}
\toprule	Setup	&	\textit{Squares}	&	MNIST-3/5	&	\textit{COPD}	&	MNIST	&	CIFAR-10 \\	\midrule
B		& 82.84$\pm$26.56		& 71.70$\pm$47.34		& 3.953$\pm$2.347		& 10.74$\pm$.89		& 25.89$\pm$2.23	\\
$L_\alpha$		& 06.91$\pm$00.87		& 12.42$\pm$00.68		& 1.424$\pm$0.680		& 05.84$\pm$.50		& 24.00$\pm$0.55	\\
P		& 09.77$\pm$08.26		& 15.04$\pm$04.14		& 0.664$\pm$0.291		& 04.01$\pm$.53		& 16.83$\pm$0.30	\\

 	\bottomrule
\end{tabular}
\end{small}
\end{center}

\end{table*}
For the robustness metric, we report the estimated point $\epsilon_{50\%}$, where 50\% of test examples are incorrectly classified after applying \gls*{PGD} attack with varying values of $\epsilon$. We report a single value instead of the usual curve of accuracy as a function of $\epsilon$ to have a more objective evaluation. We used accuracy as the basis of the metric to be comparable to the style of reporting in the literature. The defined $\epsilon_{50\%}$ is equivalent to considering misclassified inputs to have $\rho(x)=0$ and calculating the median of the estimated robustness, which has been used to evaluate the baseline metric $\overline{\alpha_{x}}$~\citep{alignment}. In addition to using PGD with $p=\infty$, we calculated the robustness of the models against $L^2$-constrained \gls*{PGD} attacks, with an adapted step size $\eta_{L^2}=\eta_{L^\infty}\times\sqrt{n}$, where $n$ is the dimensionality of the data, and against the Square Attack~\citep{blbox}, an iterative black-box attack, and the \gls*{STA}~\cite{sta}. Since the Square Attack is formulated for images, we adapted it to the \textit{Spheres} dataset by reshaping its 500-feature vectors to a 20$\times$25 image. We used the black-box attack to evaluate if any defenses were causing gradient obfuscation~\citep{obfuscated}. For the MNIST, MNIST-3/5, CIFAR-10, and the \textit{COPD} datasets, training with the alignment penalty as defined in (\ref{eq:penalty}) employed a distinct $G$ for each run, and, for calculating the reported $\overline{\alpha_{\Delta x}}$, a different $G$ than the one used for training. We used the correct $\Delta x$ to calculate $L_{\alpha}$ and $\overline{\alpha_{\Delta x}}$ for the \textit{Spheres} and \textit{Squares} datasets. Table \ref{table:results} presents results for robustness and alignment. Graphs of accuracy as a function of $\epsilon$ are given in Section \ref*{sec:graphs} in the Supplementary Material. 

The alignment $\overline{\alpha_{\Delta x}}$ increased for all \gls*{PGD}-trained models (rows P) when compared to the baseline (rows B). Similarly, the robustness of all models trained with the $L_\alpha$ penalty increased when compared to the baseline (rows B). These results show that alignment $\overline{\alpha_{\Delta x}}$ and robustness are closely related, and one is a consequence of the other. Theoretically, for the \textit{Spheres} dataset, $\rho_x/\overline{\alpha_{\Delta x}}=0.15$ when $p=2$. This value is very close to the ratios of corresponding values in Table \ref{table:results}, which lie between 0.148 and 0.151. 

Table \ref{table:resultssta} presents the results of the robustness  against an attack without a constraint on the $L^p$ norm, the \gls*{STA}~\cite{sta}, which produces adversarial attacks by applying deformation fields to images. The theoretical formulation of our alignment metric considers robustness against $L^2$ attacks. However, if the non-$L^p$ attack produces on-manifold examples, the formulation should still be valid, considering robustness is defined as the distance to the support of each class. Results are presented as $\tau_{50\%}$, the $\tau$ for which 50\% of the test examples are misclassified after adversarial attacks. The hyperparameter $\tau$ controls the strength of the total variation penalty in the \gls*{STA} formulation, controlling how locally smooth the spatial gradient of the deformation field is. The lower the $\tau_{50\%}$, the more robust the model is. Results are similar to Table \ref{table:results}, except for the robustness of the models trained with $L_{\alpha}$ penalty for the CIFAR-10 dataset, which had robustness similar to the baseline model, and for the MNIST-3/5 dataset, which produced a more robust model than PGD-training.

In Table \ref{table:results}, models trained using PGD exhibited the strongest signs of gradient obfuscation, highlighted by the black-box attack (column $\epsilon_{50\%}$, BlBox$_{p=\infty}$) being considerably more potent than the PGD attack (column $\epsilon_{50\%}$, PGD${p=\infty}$) for some datasets. Section \ref{sec:graphs} in the Supplementary Material provides an analysis of gradient obfuscation using the graphs of accuracy as a function of the perturbation norm.

Except for the \textit{Spheres} dataset, where our proposed alignment metric $\alpha_{\Delta x}$ mathematically reduces to the alignment metric $\alpha_x$~\citep{alignment}, our metric $\overline{\alpha_{\Delta x}}$ was larger than $\overline{\alpha_x}$ in all cases, demonstrating that robust models are more closely aligned with $\Delta x$ than with $x$. The alignment $\overline{\alpha_{\Delta x}}$ employs the direction to which the gradient is pointing, providing more information than the $\overline{\alpha_x}$ metric, which has an absolute value in its numerator. Furthermore, in addition to a different alignment direction ($\Delta x$), our proposed metric proposes a different definition of what is aligning to that direction, $\nabla_{\text{logit}(x)_{\tilde{c}(x)}-\text{logit}(x)_{y}}$ against $\nabla_{\text{logit}(x)_{m(x)}}$ for the baseline metric $\overline{\alpha_{x}}$. The metric we reported corresponds to the highest alignment for the baseline metric considering several possible methods for calculating the input gradient, as shown in Section \ref{sec:alignments} in the Supplementary Material. 

For most of the datasets, even though the penalty alignment training (rows $L_\alpha$) had the closest alignment (columns $\overline{\alpha_{\Delta x}}$ and $\overline{\alpha_{x}}$), \gls*{PGD} (rows P) had the highest robustness when $p=\infty$ (column $\epsilon_{50\%}$, PGD$_{p=\infty}$). \gls*{PGD} likely not only aligns the gradient but also improves robustness in other ways, such as possibly providing a denser sampling of inputs, especially in critical regions, and making the model more locally linear~\citep{linearity}. When comparing results between MNIST and MNIST-3/5, robustness was similar for the $L_\alpha$ training method, but $\overline{\alpha_{\Delta x}}$ was lower for \gls*{PGD}-training for MNIST. The $\overline{\alpha_{\Delta x}}$ achieved for the CIFAR-10 dataset showed that the alignment with $\Delta x$ is sometimes easier to learn with \gls*{PGD} than with $L_\alpha$. 

When setting the training method to PGD-training, with varying values of $\epsilon$, both metrics showed a good Pearson correlation with robustness, as seen in Table \ref{table:correlation}. Our metric showed a better correlation for most datasets. We also calculated the Pearson correlation for the ImageNet dataset~\cite{imagenet} using six models provided by the RobustBench library~\cite{robustbench}. The six models combined three different architectures and three different defense techniques. The robustness and alignment metrics values for each model are provided in Table \ref{table:robustbenchdetails}. The calculated correlations can be found in Table \ref{table:robustbench}. Despite our alignment being lower for this dataset, the correlation was higher. The more challenging generative task might cause the lower alignment. More details about how ImageNet numbers were calculated can be found in Section \ref{sec:imagenet} of the Supplementary Material. Graphs of the data used to calculate these correlations for all other datasets are provided in Section \ref{sec:correlation} of the Supplementary Material.

\begin{table}[tbh!]

\begin{center}
\caption{Pearson correlation between alignment metrics and robustness against \gls{PGD} attack ($\epsilon_{50\%}$, PGD$_{p=\infty}$) for our proposed metric and the baseline metric $\overline{\alpha_{x}}$~\cite{alignment}, with PGD-training as a constant training method. For each dataset, five models were trained for each of six or seven values of $\epsilon$.}

\label{table:correlation}
\begin{small}
\begin{tabular}{@{}ccccccc@{}}
\toprule														
Metric	&	\textit{Spheres}	&	\textit{Squares}	&	MNIST-3/5	&	\textit{COPD}	&	MNIST	&	CIFAR-10	\\	\midrule
$\overline{\alpha_{\Delta x}}$ (ours)	&	.818	&	.883	&	.861	&	.280	&	.897	&	.996	\\	
$\overline{\alpha_{x}}$ 	&	.818	&	.870	&	.591	&	.656	&	.185	&	.994	\\	\bottomrule

\end{tabular}
\end{small}
\end{center}

\end{table}

\begin{table}[tbh!]

\begin{center}
\caption{ Alignment and robustness against $PGD_{p=\infty}$ attacks for ImageNet models. The names provided in the Model column correspond to the names given to them in the Robust
bench library.}

\label{table:robustbenchdetails}
\begin{small}
\begin{tabular}{@{}cccc@{}}
								\toprule
Model	&	Robustness	&	$\overline{\alpha_{\Delta x}}$ (ours)	&	$\overline{\alpha_{x}}$	\\	\midrule
Standard\_R50	&	0.001	&	0.000	&	0.001	\\	
Salman2020Do\_R18	&	0.004	&	0.008	&	0.016	\\	
Wong2020Fast	&	0.007	&	0.004	&	0.016	\\	
Engstrom2019Robustness	&	0.015	&	0.009	&	0.015	\\	
Salman2020Do\_R50	&	0.018	&	0.008	&	0.013	\\	
Salman2020Do\_50\_2	&	0.023	&	0.009	&	0.013	\\	\bottomrule

\end{tabular}
\end{small}
\end{center}

\end{table}

\begin{table}[tbh!]

\begin{center}
\caption{Alignment and correlation values for ImageNet models.}

\label{table:robustbench}
\begin{small}
\begin{tabular}{@{}cccc@{}}
\toprule														
Metric	&	\textit{Nonrobust} &	\textit{Robust (average)}	&	\textit{Pearson correlation}	\\	\midrule
$\overline{\alpha_{\Delta x}}$ (ours)	&	.000	&	.008	&	.670	\\	
$\overline{\alpha_{x}}$ 	&	.001	&	.015	&	.426	\\	\bottomrule

\end{tabular}
\end{small}
\end{center}

\end{table}

\subsection{Validity of local linearity assumption}
\label{sec:linearity}
\begin{figure*}[h!]

\begin{center}
\centerline{\includegraphics[width=\textwidth]{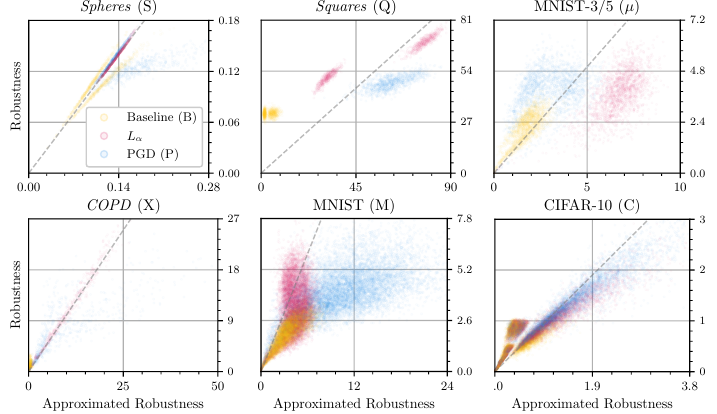}}
\caption{Robustness against $L^2$ Carlini-Wagner attack~\cite{cwattack} as a function of the approximated robustness given by Lemma \ref{lemma:locallinearity} for six datasets and three training methods. Each marker represents one test set example. The dashed diagonal gray line represents the identity function. Models were selected randomly from the five trained models for each training setup.} \label{fig:linearity}
\end{center}
\end{figure*}

\begin{table*}[tbh!]
\setlength\tabcolsep{2.7pt}
\begin{center}
\caption{Average alignment $\overline{\alpha_{\Delta x}}$ values for the two multiclass datasets and three training methods from the original metric ($\tilde{c}(x)$) and after replacing $\tilde{c}(x)$ from Equation \ref{eq:metric} with all other classes, except the ground truth class $y$ (Other).}
\label{table:alphalinearity}
\begin{small}
\begin{tabular}{@{}ccccccc@{}}
\toprule														
Class & M-B	&	M-$L_{\alpha}$	&	M-P	&	C-B	&	C-$L_\alpha$	&	C-P	\\ \midrule
$\tilde{c}(x)$ & .075 $\pm$ .006 & .592 $\pm$ .009 & .181 $\pm$ .012 & .008 $\pm$ .003 & .025 $\pm$ .002 & .041 $\pm$ .001 \\
Other & .056 $\pm$ .004	&	.478 $\pm$ .008	&	.132 $\pm$ .017	&	.008 $\pm$ .001	&	.027 $\pm$ .004	&	.038 $\pm$ .002	

\\	\bottomrule

\end{tabular}
\end{small}
\end{center}
\end{table*}
Figure \ref{fig:linearity} qualitatively surveys the validity of the equations from Lemma \ref{lemma:locallinearity} as an approximation of a model's robustness. For a minority of the datasets and methods, the approximated robustness is very similar to the robustness. The fact that the approximation from Lemma \ref{lemma:locallinearity} does not hold perfectly may be one of the reasons for an imperfect correlation between alignment and robustness. Furthermore, the metric might still present relatively good results because, in its final formulation, the main impact of the local linearity assumption is in the calculation of class $\tilde{c}(x)$. In Equation \ref{eq:metric},  $\tilde{c}(x)$ is used in both compared vectors, $\Delta x$ and $\nabla_{\ell(x)}$, probably resulting in analogous comparisons for all possible classes. We calculated the average alignment after replacing $\tilde{c}(x)$ from Equation \ref{eq:metric} with all other classes that are not the ground truth. The results from Table \ref{table:alphalinearity} show that the alignment for other classes is relatively close to the original alignment using $\tilde{c}(x)$. For binary datasets, there is always only one possibility for $\tilde{c}(x)$, so results for these datasets were not calculated.

\subsection{Qualitative analysis of the changes to the input gradient}
\label{sec:qualitative}
\begin{figure}[h]
\begin{center}
\centerline{\includegraphics[width=0.45\linewidth]{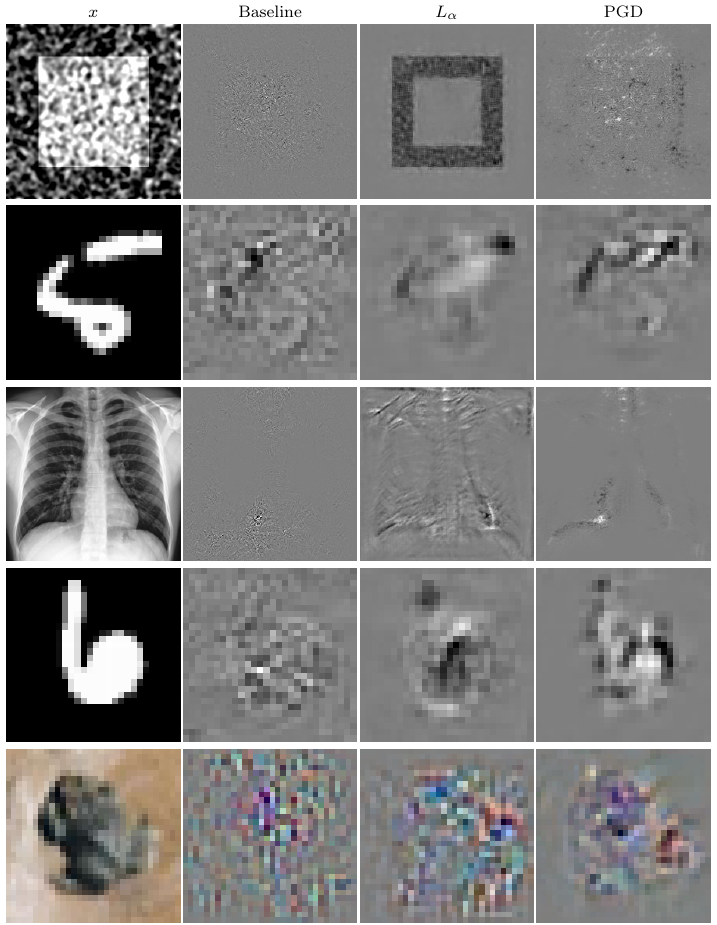}}
\caption{Examples of the calculated $\nabla_{\ell(x)}$ for all the training methods for the datasets, from top to bottom:  \textit{Squares}, MNIST-3/5, \textit{COPD}, MNIST, CIFAR-10.} \label{fig:interpretable}
\end{center}
\end{figure}
Figure \ref{fig:interpretable} shows $\nabla_{\ell(x)}$ for random images in each dataset for all three training methods. The calculated $\nabla_{\ell(x)}$ are noisier for the baseline and smoother for models trained with $L_\alpha$, whereas \gls*{PGD}-trained models have an intermediate amount of noise and are more localized. Part of the differences in the $\nabla_{\ell(x)}$ produced for the MNIST dataset for the $L_{\alpha}$ and PGD methods is caused by the different $\tilde{c}(x)$ in each case. For this specific sample, $\tilde{c}(x)=0$ for the model trained with $L_{\alpha}$ and $\tilde{c}(x)=4$ for the model trained with PGD.

\section{Conclusion}

We proposed a novel alignment direction for the gradient of robust models as the vector pointing to the closest example of the support of the closest inaccurate class. We validated the proposed direction theoretically and showed experimentally that alignment increases with \gls*{PGD} training and that robustness increases with alignment enforcement. Trained models also showed a closer alignment with the proposed metric than with another metric definition. 

The proposed metric was not directly predictive of robustness since models with the strongest alignment were not always the most robust. This finding is possibly a result of the violation of the strong assumptions of the theoretical analysis. It also highlights the possibility of \gls*{PGD} increasing robustness by means other than gradient alignment. Additionally, the proposed method for calculating $\Delta x$ for the use of the metric is complex to calculate, possibly limiting its use. Despite not being the focus of the paper, the proposed defense technique is also less strong than the baseline defense PGD.

When comparing models trained with \gls{PGD} with varying levels of robustness, the proposed metric correlated better with the robustness than the baseline metric for most datasets. Finally, we expand our fundamental understanding of adversarial defenses, benefiting future analyses of model robustness.

% The theoretical analysis is also based in assumptions that are very unlikely to hold in a real-world application. In this paper, the local linearity assumption is shown not to hold, for example. The correlation between robustness and metric still has space for improvements. 

Future work might include investigating less noisy and faster methods for generating $\Delta x$. For example, the use of CycleGAN~\cite{cyclegan} can be analyzed for the direct generation of $\widehat{\Delta x}$ as a more restrictive method for limiting the changes between classes to only needed changes. The use of the Wasserstein GAN~\cite{wgan} loss might also help improve stability. For the indirect formulation, the use of a conditional StyleGAN2~\cite{stylegan} might show some improvements in the transformation between classes by disentangling the latent space and simplifying the optimization process. Ultimately, the lack of stability in the training of GANs does not play an essential role in the potential future performance of our proposed alignment. If a stable generative method/model is proposed/trained with good results, our method can be adapted to it. Therefore, the future progress of generative methods from the literature may also facilitate the use and performance of our alignment metric. 

Another potential future work is the exploration of stronger and more complex defenses inspired by the proposed alignment metric. For example, the use of the direction perpendicular to $\widehat{\Delta x}$ when sampling examples for adversarial training, avoiding sampling in the direction of other classes, might have positive results. We might also change the defense formulation by considering that $\Delta x$ might have multiple solutions and making a more lenient alignment penalty. The penalty might be used, for example, with a stronger weighting to pixels that are changed less often by the generative methods.

% the theoretical calculation of the impact of the violated assumptions through the calculation of bounds for the relationship between alignment of gradient of a model and its robustness; 

\section*{Acknowledgements}
Copyediting support was provided by Christine Pickett. Funding: This work was supported by the National Institute Of Biomedical Imaging And Bioengineering of the National Institutes of Health [grant numbers R21EB028367]. The funding source had no other involvement in the study. 

\bibliography{references_short}
\medskip

\textbf{Ricardo Bigolin Lanfredi} received his BS degree in Electrical Engineering from the Federal University of Rio Grande do Sul, his M.Eng. degree from CentraleSupélec, and is in 2021 a PhD candidate at the University of Utah working with medical image analysis and deep learning at the Scientific Computing and Imaging Institute.

\textbf{Joyce Schroeder MD} is a Professor in the Department of Radiology and Imaging Sciences, School of Medicine, University of Utah, with subspecialty in cardiothoracic imaging. Her academic work is focused on quantitative imaging and machine learning in investigative studies including longitudinal NIH studies in chronic lung disease and cardiopulmonary processes.

\textbf{Tolga Tasdizen} is a Professor of Electrical and Computer Engineering at the University of Utah.  He earned the BS degree in EE from Bogazici University and the PhD degree in Engineering from Brown University. His research interests are in the areas of machine learning and image analysis. 
\clearpage
\newpage
\phantomsection

 \beginsupplement
 \section*{Supplementary Material - Quantifying the Preferential Direction of the Model Gradient in Adversarial Training With Projected Gradient Descent}
 \subsection{Experimental setup and dataset details}
 \label{sec:setup}
  \begin{table*}[htp]

\begin{center}
\caption{Details about datasets and training setup. ART stands for average running time (in hours), and $\%_{-1}$ for percentage of samples from class -1. The reported ART might be longer than usual runs since sometimes more than one script was run in the same GPU at the same time.}  \label{table:setup}
\begin{footnotesize}
\setlength\tabcolsep{2.5pt}
\begin{tabular}{@{}	l		c		c		c		c	c    c c @{}}	\midrule
	Properties	&	\textit{Spheres}	&	\textit{Squares}	&	MNIST-3/5	&	\textit{COPD} &	MNIST & CIFAR-10 & \textit{Squares32}	\\	\midrule
	Size Train	&	$\frac{10,000,000}{\text{epoch}}$ &	10,000	&	10,397	&	3,711 &	54,000 & 45,000 & 10,000\\	
	Size Validation	&	200	&	200	&	1,155	&	596	& 6,000 & 5,000 & 200\\	
	Size Test	&	1,000	&	1,000	&	1,902	&	950	& 10,000 & 10,000 & 1,000\\	
	Data size	&	500	&	224$\times$224	&	28$\times$28	&	224$\times$224 &	28$\times$28 &	32$\times$32$\times$3 & 32$\times$32	\\	
	$\%_{-1}$ Training	&	$\sim$50\%	&	49.7\%	&	53.1\%	&	63\% & - & - & 49.7\%	\\	
	$\%_{-1}$ Validation	&	$\sim$50\%	&	50.5\%	&	53.1\%	&	49.8\% & - & - &	50.5\%\\	
	$\%_{-1}$ Test	&	$\sim$50\%	&	48.6\%	&	53.1\%	&	58.2\% & - & -	& 48.6\% \\	
	PGD train $\epsilon$	&	0.005	&	0.2	&	0.3	&	0.006 &	0.3 & 0.01 & -\\	
	PGD step $\eta$	&	0.002	&	0.02	&	0.02	&	0.02 & 0.02 & 0.02	& -\\	
	\# epochs	&	80	&	30	&	30	&	30 & 30 & 100 & -	\\	
	$\lambda_\alpha$ ($L_{\alpha}$)	&	0.1	&	0.1	&	0.1	&	10	& 0.3 & 0.3 & -\\
	$\lambda_{Reg\hspace{0.02cm}z}$ ($L_{proj}$)	& -	&	-	&	-	&	- & 1.5 & 0.03 & 0.03\\
	Batch size	&	50	&	12	&	12	&	12 & 64 & 128 & -	\\	
	ART baseline	&	2.3	&	1.3	&	4.1	&	3.7	& 6.5 & 15.5 & -\\	
	ART GAN	&	3	&	0.8	&	0.1	&	1 & 61.7 &	54.7 & 3.2\\	
	ART $L_\alpha$	&	6.2	&	1.5	&	3.7	&	4.2	& 6.0 & 13.1 & -\\
	ART PGD	&	15.4	&	2.5	&	3.1	&	4.9	& 9.1 & 27.0 & -\\	
	ART BlBox	&	0.3	&	6.3	&	8.8	&	3.8 & 15.7 & 17.1 & -\\
	ART STA	&	-	&	4.1	&	0.3	&	2.3 & 1.1 & 2.2 & -\\
	 \bottomrule\end{tabular}
\end{footnotesize}
\end{center}

\end{table*}
We used PyTorch~\cite{pytorch} to build our experiments. Other libraries and their versions can be found on the project's code repository\footnote{\url{https://github.com/ricbl/gradient-direction-of-robust-models}}. For the \textit{Spheres} dataset, we used a 2-hidden layer network, with 1000 neurons per layer and ReLU non-linearity, and a last-layer output of a 500-dimensional vector for the generator and a scalar for the classifiers. For all other datasets, we used a Resnet-18~\citep{resnet} as the classifier, with weights pre-trained for ImageNet loaded from PyTorch. \gls*{PGD} attacks, for training and validation, used number of steps $\kappa=40$. The Square Attack was used with 5000 queries per attack and 80\% as an initial percentage of features to be modified. We used the Adam optimizer~\citep{Adam} with a learning rate of $10^{-4}$, except for the CIFAR-10 dataset. Hyperparameters were chosen by checking for best robustness given that the accuracy with no attack does not drop by more than 4\%. The best epoch was chosen by the highest robustness against \gls*{PGD} ($p=\infty$) on the validation set. For all datasets for which we found little or no impact on the accuracy, i.e., all datasets except the CIFAR-10 dataset, batch normalization~\citep{bn} was turned off to avoid its influence on robustness~\cite{bndrop}. For binary datasets, reported results used a single output for the classifier. Further details on hyperparameters specific to each dataset are given in Table \ref{table:setup}.

For the CIFAR-10 dataset, a few aspects were modified to improve accuracy:
\begin{itemize}
\item we changed the first convolutional layer of Resnet-18 to have a 3$\times$3 kernel, a stride of 1, and no max pooling immediately after it;
\item we initialized Resnet-18 convolutional weights using the Kaiming uniform initialization~\cite{kaiming} instead of ImageNet pre-trained weights;
\item we used a stochastic gradient descent optimizer, with the learning rate equals to 0.01;
\item we checked for the best epoch only after 33 epochs of training because of instabilities in the calculated validation robustness before the validation accuracy is stable;
\item we left batch normalization turned on.
\end{itemize}

The generator for the direct method for estimating $\Delta x$ used a U-net~\citep{unet} architecture. In the U-net, we utilized two levels of downsampling for MNIST-3/5 and four levels for \textit{Squares} and \textit{COPD}. For training the U-net, we used $\lambda_{G}=0.3$, $\lambda_{RegG}=0.5$, $\lambda_{Dx}=1$ and $\lambda_{D\hat{x}}=0.01$. The values of $\neg y$ and $y$ were concatenated as channels to the U-net's bottleneck. The best epoch during the U-net training was chosen by checking minimal total loss.

 The computer infrastructure employed included 11 Titan V, 6 Titan RTX, and 8 Titan V100 SMX2, and all GPUs were used interchangeably depending on availability. Some of the experiments required large GPU memory capacity, which was available only on Titan RTX. Training, combined with best epoch validation, took between 5 minutes and 62 hours for each run, depending on the dataset, method, and GPU used. The average time for each method and dataset is reported in Table \ref{table:setup}. Test evaluations for PGD attack took less than 1 hour each. Table \ref{table:setup} also presents further quantitative detail about the datasets used.

The \textit{COPD} dataset was filtered to include only samples for which the \gls*{PFT} was acquired within 30 days of the \gls*{CXR}. Patients with a \gls*{PFT} indicating the presence of \gls*{COPD} were assigned to class 1. Images were center-cropped, resized to $256\times256$, and cropped to $224\times224$ (randomly in training), and they had their histograms equalized and range adjusted to $[-1,1]$. The dataset was split into training, validation, and test sets by patient ID, since some patients were associated with more than one sample.

\subsection{Proofs}
\label{sec:proofs}

\Paste{th1}
\begin{proof}

Since we assume local linearity, the model can be represented by $logit(x)=W^Tx+b$ by using Lemma \ref{lemma:locallinearity}. Since $\{c^*(x_i),m(x_i)\}=\{i,j\}$ and $\{c^*(x_j),m(x_j)\}=\{i,j\}$, the only two classes involved in the calculation of robustness for both $x_i$ and $x_j$ are $i$ and $j$, i.e., the robustness is always measured between the input example and the decision boundary separating $i$ and $j$. We can then represent $logit(x)=W^Tx+b$ as a binary model to calculate robustness, given by $logit_b(x)=\langle w,x \rangle+b_b=logit(x)_j-logit(x)_i$, and where positive outputs are equivalent to outputting class $j$. We note that $w=W_{:,j}-W_{:,i}=\nabla{\text{logit}(x_i)_j}-\nabla{\text{logit}(x_i)_i}=\nabla{\text{logit}(x_j)_j}-\nabla{\text{logit}(x_j)_i}$ and that $b_b = b_j-b_i$. The alignment can be simplified as 
\begin{equation}
\begin{aligned}
\label{eq:alphasimpl}
\alpha=\textit{sim}(x_j-x_i,\nabla{\text{logit}(x_i)_j}-\nabla{\text{logit}(x_i)_i})=\\=\textit{sim}(x_j-x_i,w)= \frac{\langle x_j,w\rangle -\langle x_i,w\rangle }{\left\lVert x_j-x_i \right\rVert_2 \left\lVert w \right\rVert_2}.
\end{aligned}
\end{equation}
If $x$ is correctly classified, $\rho(x)$ is equal to the distance between $x$ and the decision boundary. In the case of misclassification, we use the negative of the distance. We can use the equation of signed distance between $x$ and the hyperplane defined by $\langle w,x\rangle +b_b=0$ and the result from (\ref{eq:alphasimpl}) to get

\begin{equation}
\rho(x_i) = -\frac{\langle x_i,w\rangle +b_b}{\left\lVert    w \right\rVert_2},\, \rho(x_j) = \frac{\langle x_j,w\rangle +b_b}{\left\lVert   w \right\rVert_2},
\end{equation}

\begin{equation}
\rho(x_i) + \rho(x_j) = \frac{\langle x_j,w\rangle -\langle x_i,w\rangle }{\left\lVert w \right\rVert_2}=\alpha\times\left\lVert x_j-x_i \right\rVert_2.
\end{equation}

\end{proof}

\Paste{th2}
\begin{proof}
The expected robustness of a model $m$ can be written as
\begin{equation}
\rho_m = \sum_{c=1}^{\mathcal{C}}\int_{supp_c} P(x)\rho(x) \,dx = \sum_{k=1}^{K}\int_{\mathscr{S}_k} P(x)\rho(x) \,dx.
\end{equation}

Since we can establish a bijection in each $\mathscr{S}_k$ between $supp_{i_k}$ and $supp_{j_k}$, we can integrate over both supports at the same time, pair by pair of $x_{i_k}$ and $x_{j_k}$. Since $P(x_{i_k})=P(x_{j_k})$, we can factor the probability, resulting in

\begin{equation}
\rho_m = \sum_{k=1}^{K}\int_{supp_{i_k},supp_{j_k}} P(x_{i_k})(\rho(x_{i_k})+\rho(x_{j_k})) \,dx.
\end{equation}

Sampling is balanced between the two classes in $\mathscr{S}_k$ since, from the established bijection,
\begin{equation}
\label{eq:halfprob_multi}
\int_{supp_{i_k}} P(x)\,dx = \int_{supp_{j_k}} P(x) \,dx = \frac{1}{2} \int_{\mathscr{S}_k} P(x)\,dx.
\end{equation}

Using Theorem \ref{theorem:pair} to substitute for $\rho(x_{i_k})+\rho(x_{j_k})$ and using (\ref{eq:halfprob_multi}), we reach the two inequalities given by the theorem,
\begin{equation}
\begin{split}
\rho_m = \sum_{k=1}^{K}\int_{supp_{i_k},supp_{j_k}} P(x_{i_k})\alpha_{i_k} \left\lVert x_{i_k}-x_{j_k} \right\rVert_2 \,dx \ge\\\ge \sum_{k=1}^{K}\inf(\mathscr{D}) \int_{supp_{i_k}} P(x_{i_k})\alpha_{i_k} dx= \\
= \frac{\inf(\mathscr{D})}{2} \int_{\mathscr{S}_k} P(x)\alpha dx = \frac{\inf(\mathscr{D})\times \overline{\alpha}}{2},
\end{split}
\end{equation}
\begin{equation}
\begin{split}
\rho_m = \sum_{k=1}^{K}\int_{supp_{i_k},supp_{j_k}} P(x_{i_k})\alpha_{i_k} \left\lVert x_{i_k}-x_{j_k} \right\rVert_2 \,dx \le\\\le \sum_{k=1}^{K}\sup(\mathscr{D}) \int_{supp_{i_k}} P(x_{i_k})\alpha_{i_k} dx= \\
= \frac{\sup(\mathscr{D})}{2} \int_{\mathscr{S}_k} P(x)\alpha dx= \frac{\sup(\mathscr{D})\times \overline{\alpha}}{2}, \,\, \overline{\alpha}\ge \frac{2\times\rho_m}{\sup(\mathscr{D})}.
\end{split}
\end{equation}
\end{proof}

\subsection{Details of the indirect generation of $\widehat{\Delta x}$}
\label{sec:cgan}

For the CIFAR-10 and MNIST datasets, we selected a \gls*{cGAN} proposed by \citet{cifar10gan} that offers competitive results for the CIFAR-10 dataset. The code for the \gls*{cGAN} used in the experiments was cloned from \url{https://github.com/ilyakava/BigGAN-PyTorch}. We only added support for loading the MNIST and \textit{Square32} dataset and removed the fixed random seed.

Considering the optimization to find $z^*_c$, the hyperparameters were selected from a visual check of the proximity of the images to their original class and the representation of the destination class. During its optimization, $z$ was initialized to the zero vector. The first 600 iterations were calculated with $\lambda_{Reg\hspace{0.02cm}z}=0$ to facilitate the optimization process. The following 150 iterations were calculated with $\lambda_{Reg\hspace{0.02cm}z}\ne 0$. This two-step optimization process allowed for using a single optimization per pair $(x_k,c)$, instead of several random initializations for $z$ to avoid local minima. Optimization over $z$ was performed using the Adam optimizer~\citep{Adam}, with a learning rate equals to $0.1$ for CIFAR-10 and $0.2$ for MNIST and \textit{Square32}.

\subsection{Additional images generated for gradient estimation}

Figures \ref{fig:genextrasquare} to \ref{fig:genextracifar} show a set of examples for the results of the methods proposed in Section \ref{sec:gendeltax}.

\label{sec:severalganimages}
\newlength{\sizecifarimagesextra} 
\setlength{\sizecifarimagesextra}{0.057\textwidth}

\newlength{\sizebinaryimagesextra} 
\setlength{\sizebinaryimagesextra}{0.083\textwidth}

\begin{figure*}[tbh!]
\begin{center}
\centerline{\includegraphics[width=11\sizebinaryimagesextra]{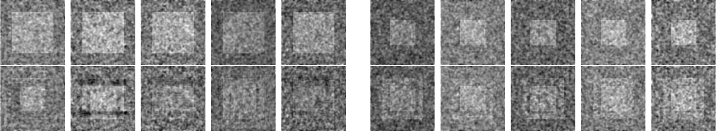}}
\caption{Examples of generated $\hat{x}^\prime$ for the \textit{Squares} dataset through the direct method. The top row coontains the original image $x$, and the bottom row contains the generated $\hat{x}^\prime$. There are five columns per class of $x$.} \label{fig:genextrasquare}
\end{center}
\end{figure*}
\begin{figure*}[tbh!]
\begin{center}
\centerline{\includegraphics[width=11\sizebinaryimagesextra]{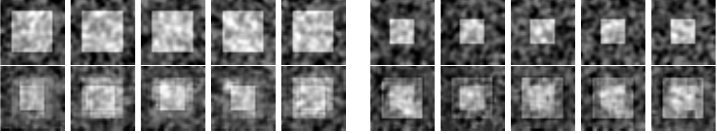}}
\caption{Examples of generated $\hat{x}^\prime$ for the \textit{Squares32} dataset through the indirect method. The top row contains the original image $x$, and the bottom row contains the generated $\hat{x}^\prime$. There are five columns per class of $x$.} \label{fig:genextrasquare32}
\end{center}
\end{figure*}
\begin{figure*}[tbh!]
\begin{center}
\centerline{\includegraphics[width=11\sizebinaryimagesextra]{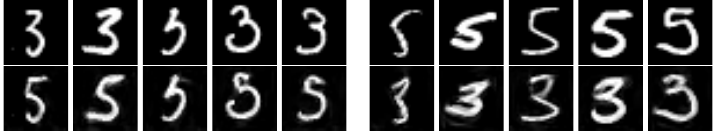}}
\caption{Examples of generated $\hat{x}^\prime$ for the MNIST-3/5 dataset through the direct method. The top row contains the original image $x$, and the bottom row contains the generated $\hat{x}^\prime$. There are five columns per class of $x$.} \label{fig:genextramnist35}
\end{center}
\end{figure*}
\begin{figure*}[tbh!]
\begin{center}
\centerline{\includegraphics[width=11\sizebinaryimagesextra]{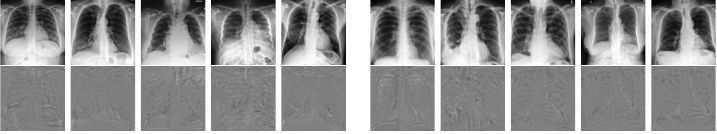}}
\caption{Examples of generated $\hat{x}^\prime$ for the \textit{COPD} dataset through the direct method. The top row contains the original image $x$, and the bottom row contains the generated $\widehat{\Delta x}$. There are five grouped columns per class of $x$, from left to right: class -1, class 1. We show $\widehat{\Delta x}$ because changes are small and difficult to perceive in $\hat{x}^\prime$. } \label{fig:genextracopd}
\end{center}
\end{figure*}
\begin{figure*}[h!]
\begin{center}
\centerline{\includegraphics[width=12\sizecifarimagesextra]{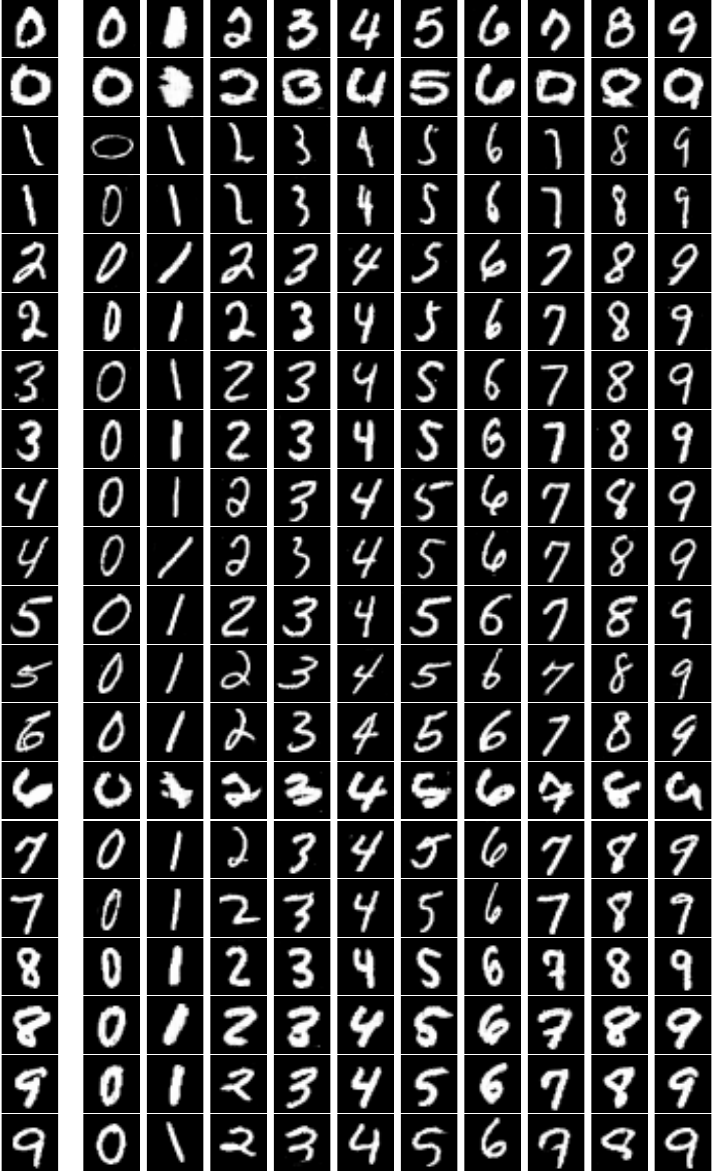}}
\caption{Examples of generated $\hat{x}^\prime$ for the MNIST dataset through the indirect method. There are two rows per class of $x$ (left column), and 10 columns on the right of $x$ to represent each of the 10 destination classes.} \label{fig:genextramnist}
\end{center}
\end{figure*}
\begin{figure*}[h!]
\begin{center}
\centerline{\includegraphics[width=12\sizecifarimagesextra]{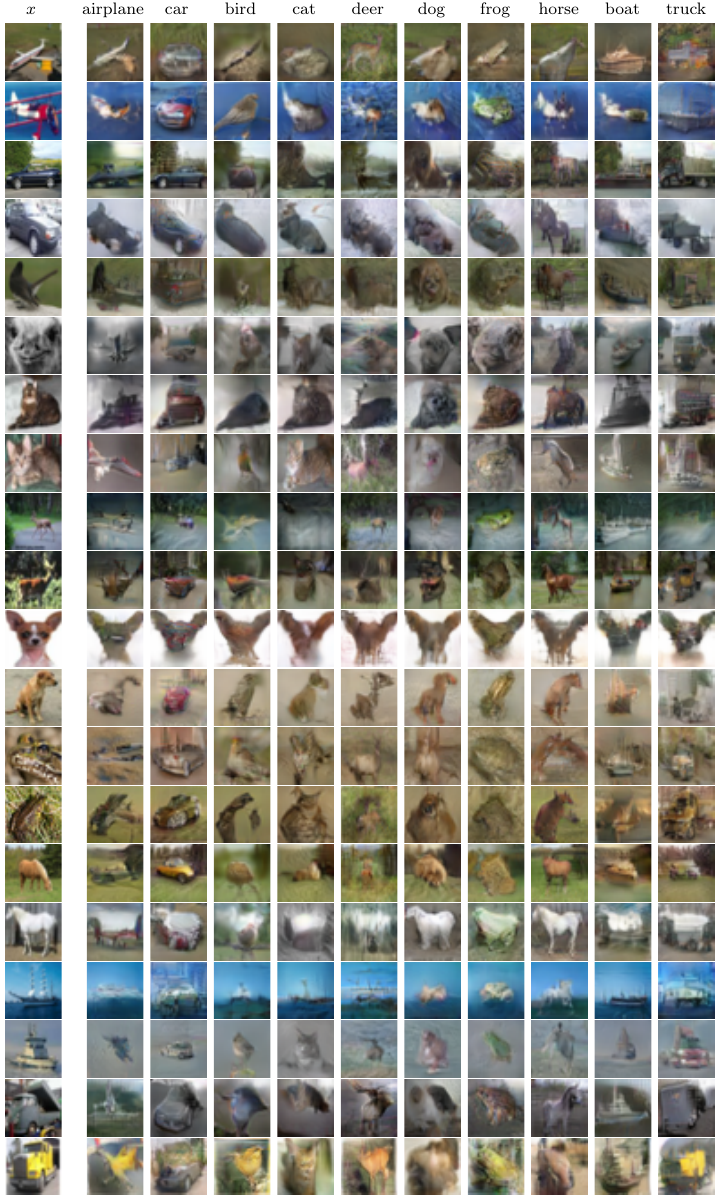}}
\caption{Examples of generated $\hat{x}^\prime$ for the CIFAR-10 dataset through the indirect method. There are two rows per class of $x$ (left column), and 10 columns on the right of $x$ to represent each of the 10 destination classes. Classes are, from top to bottom and left to right: airplane, car, bird, cat, deer, dog, frog, horse, boat, truck.} \label{fig:genextracifar}
\end{center}
\end{figure*}

\subsection{Comparison of baseline alignment metric with several input gradients}
\label{sec:alignments}

Table \ref{table:alignments} shows that the provided baseline metric corresponds to the highest alignment value when comparing several methods of calculating the input gradient.

\begin{table}[tb]

\begin{center}
\caption{Baseline alignment metric ($\overline{\alpha_x}$) values when considering distinct functions in the gradient ($\Psi\coloneqq\text{logit}(x)$). In bold, the metric we reported, coinciding with the highest values.} \label{table:alignments}
\begin{small}
\begin{tabular}{@{}lcccc@{}}
\toprule	Setup & $\Psi_{\tilde{c}}-\Psi_{y}$ & $\Psi_{c^*}-\Psi_{m(x)}$ & $\Psi_{y}$  & \boldmath$\Psi_{m(x)}$\\						\midrule
M-B	&	0.036	&	0.036	&	0.038	&	\textbf{0.038}	\\
M-$L_\alpha$	&	0.267	&	0.267	&	0.320	&	\textbf{0.320}	\\
M-P	&	0.030	&	0.030	&	0.043	&	\textbf{0.043}	\\
C-B	&	0.008	&	0.008	&	0.008	&	\textbf{0.008}	\\
C-$L_\alpha$	&	0.018	&	0.018	&	0.019	&	\textbf{0.020}	\\
C-P	&	0.029	&	0.029	&	0.031	&	\textbf{0.031}
	\\\bottomrule
\end{tabular}
\end{small}
\end{center}

\end{table}

\subsection{Correlation between alignment and robustness for a fixed training method}
\label{sec:correlation}
\begin{figure}[h]

\begin{center}
\centerline{\includegraphics[width=0.97\textwidth]{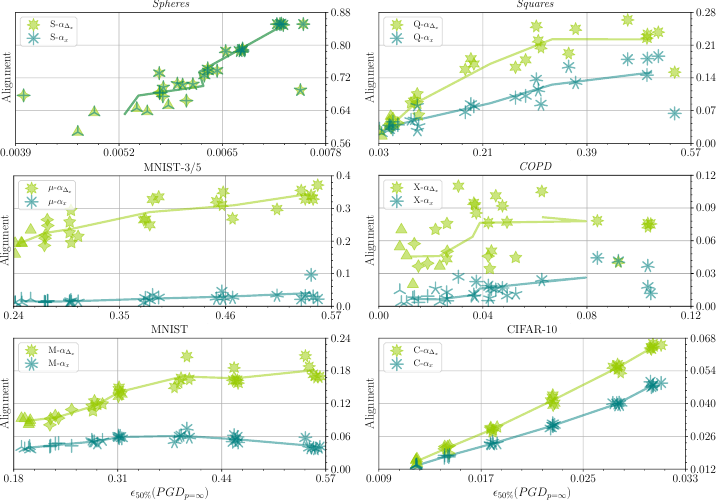}}
\caption{Alignment vs. $L_\infty$ robustness for PGD-trained models. Star-like symbols represent our proposed alignment, and asterisk-like symbols represent baseline alignments. For each dataset, the higher the number of spikes/points in a symbol, the higher the value of $\epsilon$ used for PGD-training. There are five models for each $\epsilon$. The line connects the average coordinates of each group of 5 models with fixed $\epsilon$, ordered by $\epsilon$.} \label{fig:correlation}
\end{center}
\end{figure}

Figure \ref{fig:correlation} shows alignments for a fixed training method (\gls{PGD}) and varying robustness, controlled by $\epsilon$, for six datasets.

\subsection{ImageNet validation details}
\label{sec:imagenet}
\begin{figure*}[h!]
\begin{center}
\centering
\includegraphics[width=\columnwidth]{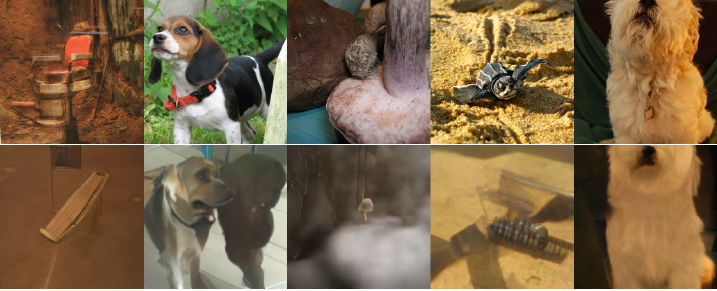}
\caption{Examples of generated $\hat{x}^\prime$ for the ImageNet dataset through the indirect method. The top row shows images from the validation set of the dataset, and the bottom row the generated images of the closest class. Classes are, from left to right: barber chair to lawn mower, beagle to Walker Hound, bolete to mushroom, leatherback turtle to screw, Lhasa Apso to Japanese Spaniel.
} \label{fig:genextraimagenet}
\end{center}
\end{figure*}

We used only 5,000 samples from the validation set for the numbers reported in this paper, following the list of samples used in the evaluation of the RobustBench leaderboard~\cite{robustbenchleaderboard}. Our generator originated from the pytorch-pretrained-biggan library \footnote{https://github.com/huggingface/pytorch-pretrained-BigGAN}, containing pretrained weights for a conditional BigGAN~\cite{biggan}. We used a learning rate equals to $0.1$ for optimizing the indirect generation of $\widehat{\Delta x}$, and $\lambda_{Reg\hspace{0.02cm}z}$ equals $0.03$. A few random images generated for the ImageNet dataset are presented in Figure \ref{fig:genextraimagenet}. %Figure \ref{fig:imagenetcorrelation} shows alignments for models from the RobustBench library for the ImageNet dataset. 

% \begin{figure}[h]

% \begin{center}
% \centerline{\includegraphics[width=0.5\textwidth]{images/correlation/imagenet_pgd_corr.pdf}}
%  \caption{Alignment vs. robustness against $PGD_{p=\infty}$ attacks for models from the RobustBench library for the ImageNet dataset. Star-like symbols represent our proposed alignment, and asterisk-like symbols represent baseline alignments. From lower $\epsilon_{50\%}$ to higher, the used models were ``Standard\_R50'', ``Salman2020Do\_R18'', ``Wong2020Fast'', ``Engstrom2019Robustness'', ``Salman2020Do\_R50'', ``Salman2020Do\_50\_2''.} \label{fig:imagenetcorrelation}
% \end{center}
% \end{figure}

\subsection{Robustness graphs}
\label{sec:graphs}
\begin{figure}[h]
\begin{center}
\centerline{\includegraphics[width=\textwidth]{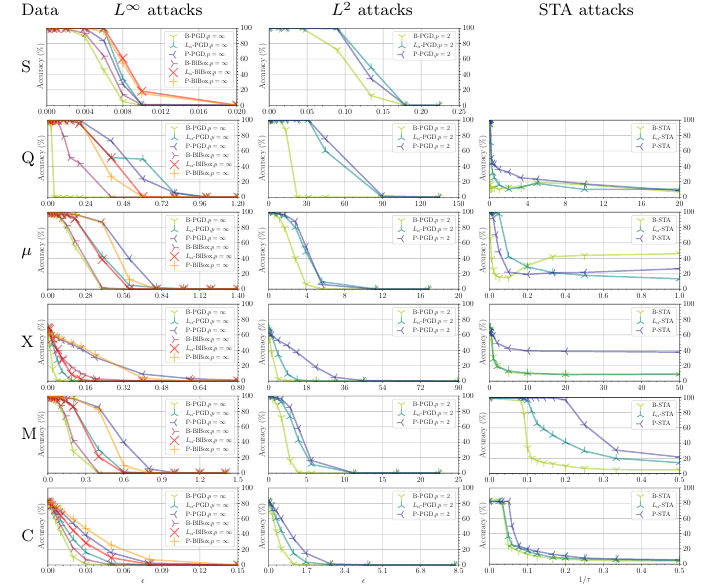}}
\caption{Accuracy of models as a function of attack strength after training using a plain supervised training baseline (B), alignment penalty ($L_{\alpha}$), and adversarial training with \gls*{PGD} (P) for six datasets: \textit{Spheres} (S), \textit{Squares} (Q), MNIST-3/5 ($\mu$), \textit{COPD} (X), MNIST (M), CIFAR-10 (C). We report results for four attacks: PGD~\citep{pgd} restricted by the $L^\infty$ norm, the black-box Square Attack ~\citep{blbox} (BlBox) restricted by the $L^\infty$ norm,  PGD~\citep{pgd} restricted by the $L^2$ norm, and Spatially Transform Attack (STA)~\cite{sta}. For attacks with a constraint on an $L^p$ norm, attack strength is measured in perturbation norm $\epsilon$, whereas for the STA attack it is shown as $1/\tau$. Models were selected randomly from the five trained models for each training setup.} \label{fig:graphs}
\end{center}
\end{figure}
In Figure \ref{fig:graphs}, all PGD attacks with a large enough bound were able to get 100\% success, and increasing the perturbation norm $\epsilon$ increased attack success rate, signs that the gradient does not suffer from intensive gradient obfuscation in any of the methods. For the \textit{Squares} dataset, the alignment penalty training method showed some gradient obfuscation for one of the classes, as seen in the bottom gap between the black-box attack and PGD attack in Figure \ref{fig:graphs}, without largely reflecting on the numbers of Table \ref{table:results}. 

\end{document}